\documentclass[10pt,twocolumn,letterpaper]{article}

\usepackage{wacv}
\usepackage{times}
\usepackage{epsfig}

\def\wacvPaperID{371} 

\wacvfinalcopy 

\ifwacvfinal
\def\assignedStartPage{9876} 
\fi

\usepackage{graphicx}
\usepackage{comment}
\usepackage{amsmath, amsthm, amssymb}
\usepackage{color}
\usepackage{cite}
\usepackage{multirow, booktabs}
\usepackage{subfig}
\newcommand{\secref}[1]{Section~\ref{#1}}
\newcommand{\figref}[1]{Figure~\ref{#1}}

\newcommand\ignore[1]{}

\usepackage{wrapfig, lipsum}

\ifwacvfinal
\usepackage[breaklinks=true,bookmarks=false]{hyperref}
\else
\usepackage[pagebackref=true,breaklinks=true,colorlinks,bookmarks=false]{hyperref}
\fi

\ifwacvfinal
\else
\fi

\begin{document}

\title{Self-Supervised Learning for Domain Adaptation on Point Clouds}

\author{Idan Achituve\\
Bar-Ilan University\\
Ramat Gan, Israel\\
{\tt\small idan.achituve@biu.ac.il}
\and
Haggai Maron\\
NVIDIA\\
Tel-Aviv, Israel\\
{\tt\small hmaron@nvidia.com}

\and
Gal Chechik\\
Bar-Ilan University, Israel\\
NVIDIA, Israel\\

}

\maketitle

\begin{abstract}
Self-supervised learning (SSL) is a technique for learning useful representations from unlabeled data. It has been applied effectively to domain adaptation (DA) on images and videos. It is still unknown if and how it can be leveraged for domain adaptation in 3D perception problems. Here we describe the first study of SSL for DA on point clouds. We introduce a new family of pretext tasks, \textit{Deformation Reconstruction}, inspired by the deformations encountered in sim-to-real transformations. In addition,  we propose a novel training procedure for labeled point cloud data motivated by the MixUp method called Point cloud Mixup (PCM). Evaluations on domain adaptations datasets for classification and segmentation, demonstrate a large improvement over existing and baseline methods.
\end{abstract}

\section{Introduction} \label{s:intro}
Self-supervised learning (SSL) was recently shown to be very effective for learning useful representations from unlabeled images \cite{dosovitskiy2015discriminative, doersch2015unsupervised, noroozi2016unsupervised, pathak2016context, gidaris2018unsupervised} or videos \cite{wang2015unsupervised, misra2016shuffle, fernando2017self, wei2018learning}. The key idea is to define an auxiliary, ``pretext" task, train using supervised techniques, and then use the learned representation for the main task of interest. While SSL is often effective for images and videos, it is still not fully understood how to apply it to other types of data. Recently, there have been some attempts at designing SSL pretext tasks for point cloud data for representation learning \cite{sauder2019self, thabet2019mortonnet, hassani2019unsupervised, zhang2019unsupervised}, yet this area of research is still largely unexplored. Since SSL operates on unlabeled data, it is natural to test its effectiveness for unsupervised domain adaptation (UDA).

Domain Adaptation (DA) has attracted significant attention recently  \cite{tzeng2014deep, ganin2015unsupervised, tzeng2017adversarial, saito2018maximum}. In UDA, one aims to classify data from a \textit{Target} distribution, but the only labeled samples available are from another, \textit{Source}, distribution. This learning setup has numerous applications, including ``sim-to-real", where a model is trained on simulated data in which labels are abundant and is tested on real-world data. Recently, SSL was successfully used in learning across domains \cite{ren2018cross, carlucci2019domain, feng2019self} and in domain adaptation for visual tasks such as object recognition and segmentation \cite{sun2019unsupervised, xu2019self}. While SSL has been used to adapt to new domains in images, it is unknown if and how SSL applies to DA for other data types, particularly for 3D data. 

The current paper addresses the challenge of developing SSL for point clouds in the context of DA. We describe an SSL approach for adapting to new point cloud distributions. Our approach is based on a multi-task architecture with a multi-head network. One head is trained using a classification or segmentation loss over the source domain, while a second head is trained using a new SSL loss.

\begin{figure}[t]
    \centering
    \includegraphics[width=0.4\textwidth]{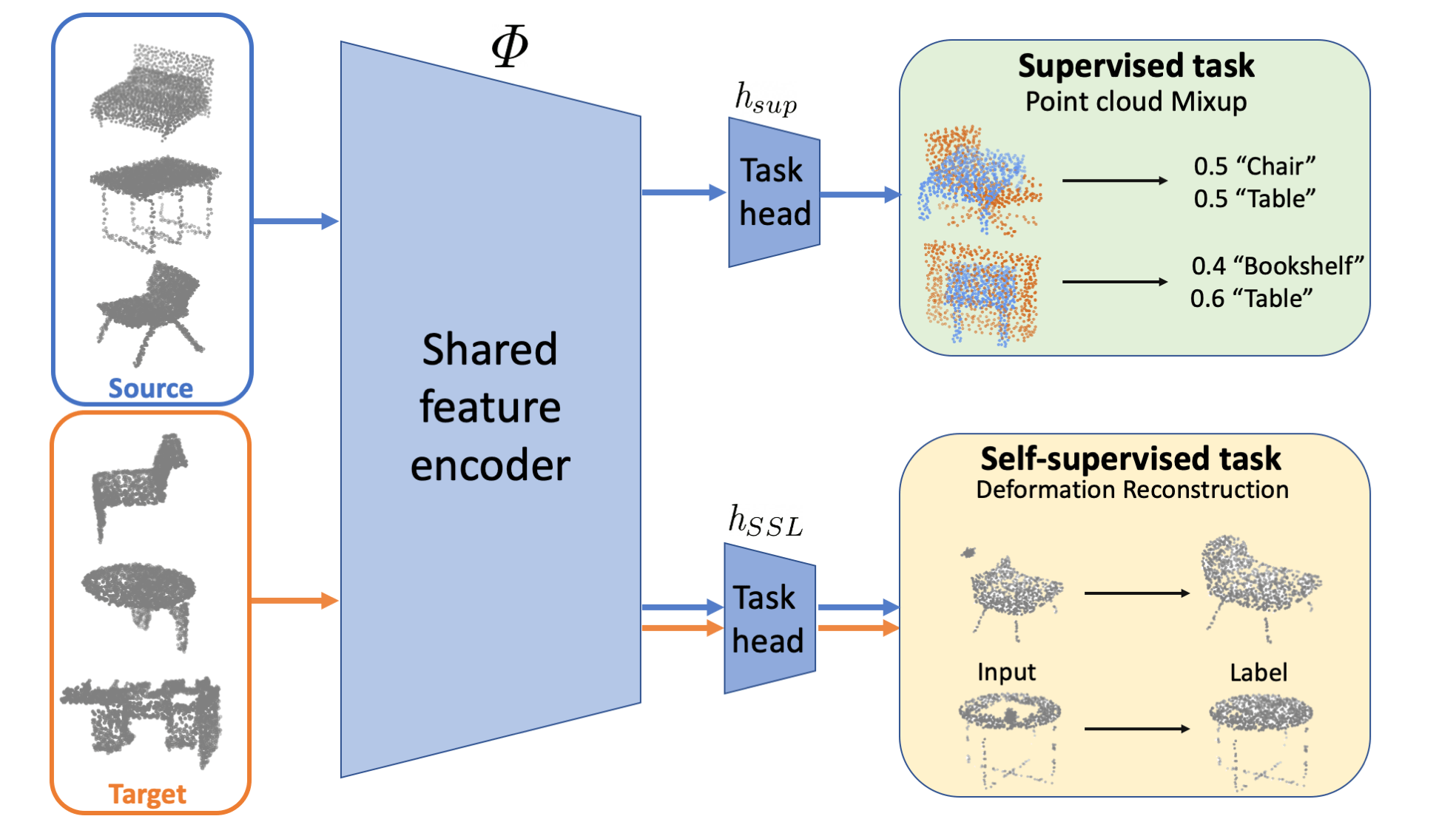}
    \caption{Adapting from a source domain of point clouds to a target domain with a different distribution. Our architecture is composed of a shared feature encoder $\Phi$, and two task-specific heads: One for the supervised task on source domain ($h_{\text{sup}}$), and another for the self-supervised task that can be applied to both domains ($h_{\text{SSL}})$.}
    \label{fig:arch}
\end{figure}

To learn a representation that captures the structure of the target domain, we develop a new family of pretext-tasks, called \textit{Deformation Reconstruction} (DefRec). We design it to address common deformations that are encountered in scanned point clouds. Scanning objects in their natural environments often leads to missing parts of the objects due to occlusion (see \figref{fig:datasets}, third column). The key idea behind the new pretext tasks is as follows: It deforms a region of the shape by dislocating some of the points; then, the network has to map back those points to their original location, reconstructing the missing regions of the shape. Importantly, success in this task requires the network to learn the underlying statistical structures of objects.

In this paper, we provide an extensive study of different approaches to deform a shape. We group these approaches into three types: (1) \textit{Volume-based deformations:} selecting a region based on proximity in the input space $\mathbb{R}^3$, (2) \textit{Feature-based deformations:} selecting regions that are semantically similar by leveraging deep point embeddings; and (3) \textit{Sampling-based deformations:} selecting a region based on three simple sampling schemes.

As a separate contribution, we propose a training procedure for labeled point cloud data motivated by the MixUp method \cite{zhang2018mixup}, called \textit{Point Cloud Mixup} (PCM). PCM is applied to source objects during training instead of the standard classification task. Together with DefRec, PCM yields large improvements over the SoTA of domain adaptation in a benchmark classification dataset in this area \cite{qin2019pointdan}. 

Finally, we designed a new DA benchmark for point cloud \textit{segmentation} based on a dataset published by \cite{maron2017convolutional}. We show that DefRec can be extended easily to segmentation tasks, leading to improved performance compared to baseline methods.

This paper makes the following novel contributions. (1) This is the first paper that studies SSL for domain adaptation on point clouds. (2) We describe DefRec, a new family of pretext tasks for point cloud data, motivated by the type of distortions encountered in sim-to-real scenario. (3) We achieve a new SoTA performance for domain adaption on point clouds, including a large improvement over previous approaches in a sim-to-real tasks. (4) We develop a new variant of the Mixup method for point cloud data. (5) A new DA benchmark for point cloud segmentation.

\section{Related work}\label{s:related}
\textbf{Deep learning on point clouds.} 
Following the success of deep neural networks on images, powerful deep architectures for learning with 3D point clouds were designed. Early methods, such as \cite{maturana2015voxnet, wu20153d, qi2016volumetric}, applied volumetric convolutions to occupancy grids generated from point clouds. These methods suffer from limited performance due to the low resolution of the discretization of 3D data. The seminal work of \cite{qi2017pointnet,zaheer2017deep} described the first models that work directly on a point cloud representation. Following these studies, a plethora of architectures was suggested, aiming to generalize convolutions to point clouds \cite{klokov2017escape, hua2018pointwise, wang2019dynamic, atzmon2018point, li2018pointcnn, su2018splatnet}. We refer the readers to a recent survey \cite{guo2019deep} for more details.

\textbf{Self-supervised learning for point clouds.}
Recently, several studies suggested using self-supervised tasks for learning meaningful representations of point cloud data, mostly as a pre-training step. In \cite{sauder2019self}, it is suggested to generate new point clouds by splitting a shape into $3\times 3\times 3$ voxels and shuffling them. The task is to find the voxel assignment that reconstructs the original point cloud. \cite{thabet2019mortonnet} proposed a network that predicts the next point in a space-filling sequence of points that covers a point cloud. \cite{zhang2019unsupervised} generated pairs of half shapes and proposed to learn a classifier to decide whether these two halves originate from the same point cloud. \cite{hassani2019unsupervised} advocates combining three tasks: clustering, prediction, and point cloud reconstruction from noisy input.  \cite{chen2019deep} learns a point cloud auto-encoder that also predicts pairwise relations between the points. \cite{tang2020improving} suggested learning local geometric properties by training a network to predict the point normal vector and curvature. In a concurrent work, \cite{alliegro2020joint} leveraged the SSL task proposed by \cite{sauder2019self}, as an auxiliary task for learning a variety of point cloud tasks. 
Compared to these studies, our work provides a systematic study of point cloud reconstruction pretext tasks specifically for DA on point clouds, a setup that was not addressed by any of the studies mentioned above.

\textbf{Domain adaptation for point clouds.}
PointDAN \cite{qin2019pointdan} designed a dataset based on three widely-used point cloud datasets: ShapeNet \cite{chang2015shapenet}, ModelNet \cite{wu20153d} and ScanNet \cite{dai2017scannet}. They proposed a model that jointly aligns local and global point cloud features for classification. \cite{su2020adapting} proposed a generic module to embed information from different domains in a shared space for object detection. Several other studies considered domain adaptation for LiDAR data with methods that do not operate directly on the unordered set of points \cite{jaritz2020xmuda, rist2019cross, wu2019squeezesegv2, saleh2019domain, yi2020complete}. \cite{jaritz2020xmuda} and \cite{yi2020complete} suggested DA methods for point cloud segmentation from a sparse voxel representation. \cite{jaritz2020xmuda} requires a paring of the point cloud and image representation of the scenes, and \cite{yi2020complete} suggested to apply segmentation on recovered 3D surfaces from the point clouds. \cite{rist2019cross} suggested a method for DA on voxelized points input using an object region proposal loss, point segmentation loss, and object regression loss. \cite{saleh2019domain} addressed the task of vehicle detection from a bird's eye view (BEV) using a CycleGAN. \cite{wu2019squeezesegv2} designed a training procedure for object segmentation of shapes projected onto a spherical surface.

\textbf{Self-supervised learning for domain adaptation.} SSL for domain adaptation is a relatively new research topic. Existing literature is mostly very recent, and is applied to the image domain, which is fundamentally different from unordered point clouds. \cite{ghifary2016deep} suggested using a shared encoder for both source and target samples followed by a classification network for source samples and a reconstruction network for target samples. \cite{xu2019self} suggested using SSL pretext tasks like image rotation and patch location prediction over a feature extractor. \cite{sun2019unsupervised} extended the solution to a multi-task problem with several SSL pretext tasks. 
\cite{carlucci2019domain} advocated the use of a Jigsaw puzzle \cite{noroozi2016unsupervised} pretext task for domain generalization and adaptation. Our approach is similar to these approaches in the basic architectural design, yet it is different in the type of data and pretext tasks.
\cite{saito2020universal} addressed the problem of universal domain adaptation by learning to cluster target data in an unsupervised manner based on labeled source data. Several other studies have shown promising results in learning useful representations via SSL for cross-domain learning. \cite{ren2018cross} suggested to train a network with synthetic data using easy-to-obtain labels for synthetic images, such as the surface normal, depth and instance contour. \cite{feng2019self} proposed using SSL pre-text tasks, such as image rotations, as part of their architecture for domain generalization.

\textbf{Deep learning of point cloud reconstruction and completion.} Numerous methods were suggested for point cloud completion and reconstruction. Most of these studies focus on high-quality shape reconstruction and completion. Our paper draws inspiration from these studies and suggests effective pretext reconstruction tasks for domain adaptation. \cite{achlioptas2018learning} suggested to learn point clouds representations with an Autoencoder (AE) based on the architecture proposed in \cite{qi2017pointnet}. In  \secref{s:results} we show that our method compares favorably to theirs. \cite{chen2019unpaired} proposed to reconstruct point clouds by training a GAN on a latent space of unpaired clean and partial point clouds. Training GANs may be challenging because of common pitfalls such as mode collapse, our method, on the other hand, is much easier to train. \cite{yu2018pu} suggested architecture for up-sampling a point cloud by learning point features and replicating them. \cite{yuan2018pcn} suggested an object completion network of partial point clouds from a global feature vector representation. \cite{wang2020cascaded} extended \cite{yuan2018pcn} approach by a cascaded refinement of the reconstructed shape.

\section{Approach}\label{s:approach}
In this section, we present the main building blocks of our approach. We first describe our general pipeline and then explain in detail our main contribution, namely, DefRec, a family of SSL tasks. We conclude the section by describing PCM, a training procedure inspired by the Mixup method\cite{zhang2018mixup} that we found to be effective when combined with DefRec. For clarity, we describe DefRec in the context of a classification task. An extension of DefRec to segmentation is detailed in \secref{sec:DefRec_for_seg}.\footnote{Code available at https://github.com/IdanAchituve/DefRec\_and\_PCM}

\subsection{Overview}
We tackle unsupervised domain adaptation for point clouds. Here, we are given labeled instances from a source distribution and unlabeled instances from a different, target, distribution. Both distributions of point clouds are based on objects labeled by the same set of classes. Given instances from both distributions, the goal is to train a model that correctly classifies samples from the target domain. 

We follow a common approach to tackle this learning setup for DA, learning a shared feature encoder, trained on two tasks \cite{xu2019d}.  (1) A supervised task on the source domain; and (2) A self-supervised task that can be trained on both source and target domains. To this end, we propose a new family of self-supervised tasks. In our self-supervised tasks, termed \textit{Deformation Reconstruction} (DefRec), we first deform a region/s in an input point cloud and then train our model to reconstruct it.

More formally, let $\mathcal{X},\mathcal{Y}$ denote our input space and label space accordingly. Let $S\subset \mathcal{X} \times \mathcal{Y}$ represent labeled data from the source domain, and $T\subset \mathcal{X}$ represent unlabeled data from the target domain. We denote by $x \in \mathbb{R} ^{n\times 3}$ the input point cloud and $\hat{x}\in \mathbb{R} ^{n\times 3}$ the deformed version of it, where $n$ is the number of points. 
Our training scheme has two separate data flows trained in an alternating fashion and in an end-to-end manner. Supervised data flow and self-supervised data flow. Both data flows use the same feature encoder $\Phi$ which is modeled by a neural network for point clouds. After being processed by the shared feature encoder, labeled source samples are further processed by a fully connected sub-network (head) denoted by $h_{\text{sup}}$ and a supervised loss is applied to their result (either the regular cross-entropy loss or a mixup variant that will be described in  \secref{s:pcm}). Similarly, after the shared feature encoder, the unlabeled source/target samples are fed into a different head, denoted $h_{\text{SSL}}$ which is in charge of producing a reconstructed version of $\widehat{x}$. A reconstruction loss is then applied to the result as we explain in the next subsection. The full architecture is depicted in Figure~\ref{fig:arch}.

\begin{figure}[!t]
    \centering
    \includegraphics[width=0.35\textwidth]{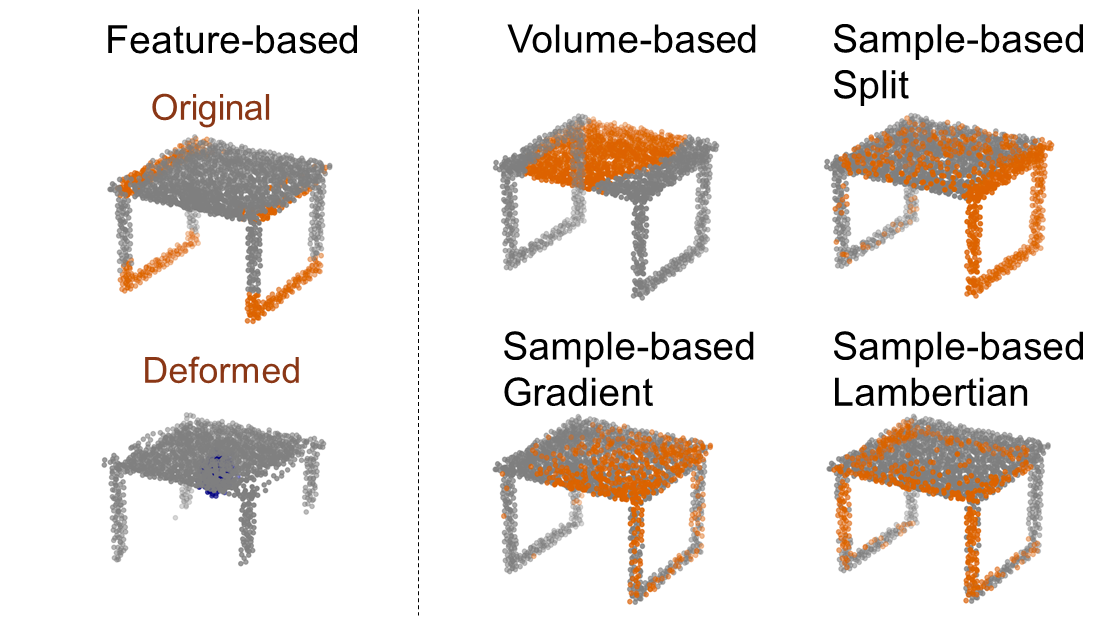}
    \caption{Illustration of five deformations of the same table object. Each method selects points to be deformed, marked in orange. Once selected, points are dislocated to a random position near the center. For the Feature-based deformation we also show the shape after dislocating the points (dislocated points are in blue).}
    \label{fig:deform_methods}
\end{figure}

\subsection{Deformation reconstruction} \label{s:DefRec}
When designing a self-supervision task, several considerations should be taken into account. First, the task should encourage the model to capture the semantic properties of the inputs. The scale of these properties is important: a task that depends on local features may not capture the semantics, and a task that depends on full global features may be over permissive. In general it is useful to focus on mesoscale features, capturing information at the scale of ``regions" or parts.

Second, for the specific case of designing SSL for DA, we want the SSL task to ``bridge" the distribution gap from the \textit{Source} to the \textit{Target} distribution. Intuitively, it would be beneficial if the SSL deformation of target samples can imitate the same deformations that are observed from source to target because then the learned representation tends to be invariant to these gaps. We designed DefRec, our family of SSL tasks, with these intuitions in mind. 


The main idea of our SSL family of tasks is to reconstruct deformed input samples. However, a key question remains: which deformations and  reconstruction tasks would produce meaningful representations for domain adaptation?  We examine three types of region selection methods: \textit{Volume-based}, \textit{Feature-based} and \textit{Sampling-based}. In all cases, the deformation is achieved by selecting a subset of points and deforming them by sampling new points from an isotropic Gaussian distribution with a small standard deviation. \figref{fig:deform_methods} illustrates all types of deformations.

\textbf{Volume-based deformations}. Perhaps the simplest and most intuitive way to define a region is based on proximity in the input space. We propose two alternatives to generate distorted point clouds.  (1) split the input space (say, the box which bounds all points in the cloud) to $k\times k\times k$ equally-sized voxels and pick one voxel $v$ uniformly at random (we found that $k=3$ works well). (2) the deformation region is a sphere with a fixed radius $r$ that is centered around a single data point $p$ selected at random. 
For both alternatives, we replace all the points with new points sampled from a Gaussian distribution around the center of $v$ (for the first method) or $p$ (for the second method).

\textbf{Feature-based deformations}. Going beyond input-space proximity, we wish to deform regions defined by their semantics.
We follow \cite{wang2019dynamic}, which showed that distances of point features taken from deeper layers capture semantic similarity more than input-space proximity. Given an input point cloud $x \in \mathbb{R} ^{n\times 3}$ we obtain a representation of the points $\Phi^l(x)$ at layer $l$, pick one point uniformly at random and based on the representations take its $k$ nearest neighbors. We replace all selected points with points sampled around the origin.

\textbf{Sample-based deformations}. 
In this case, a region is defined based on points sampled according to three common sampling protocols inspired by \cite{hermosilla2018monte}: (1) \textit{Split:} Randomly selecting a hyperplane that traverses the shape and separate it into two half-spaces. All points from the smaller part are taken, and points from the second part are randomly sampled with probability $p$ that is drawn from a uniform distribution on $[0,1]$ for each input; (2) \textit{Gradient}: Sampling points with a likelihood that decreases linearly along the largest axis of the shape; and (3) \textit{Lambertian}: A sampling method that depends on the normal orientation. For each input, we fix a “view” direction (drawn uniformly at random). The probability of sampling a point is proportional to the clamped inner product between the surface normal (which is estimated based on neighboring points) and the fixed “view” direction. In all methods, we limit the number of sampled points to be smaller than a constant to prevent large deformations. Sampled points are relocated and scattered around the origin.

\textbf{Data flow and loss function}. The self-supervised data flow starts with generating a new input-label pair $(\widehat{x},x)\in \widehat{S\cup T}\subset \mathcal{X}\times \mathcal{X}$ by using any of the methods suggested above. The deformed input $\widehat{x}$ is first processed by $\Phi$, producing a representation $\Phi(\widehat{x})$. This representation is then fed into $h_{\text{SSL}}$ which is in charge of producing a reconstructed version of $\widehat{x}$. A reconstruction loss $L_{\text{SSL}}$, which penalizes deviations between the output $h_{\text{SSL}}(\Phi(\widehat{x}))$ and the original point cloud $x$ is then applied.

We chose the loss function $L_{\text{SSL}}$ to be the Chamfer distance between the set of points in $x$ that falls in the deformed region $R$ and their corresponding outputs. More explicitly, if $I\subset \{1,\dots,n \}$ represents the indices of the points in $x\cap R$, the loss takes the following form:
\begin{equation}
\begin{aligned}
L_{SSL}(\widehat{S\cup T};\Phi,h_{\text{SSL}}) &=  \\
& \hspace{-2.0cm} \sum_{(\widehat{x},x)\in \widehat{S\cup T}} d_\text{Chamfer}\left(\{x_i\}_{i\in I},\{h_{\text{SSL}}(\Phi(\widehat{x}))_i\}_{i\in I}\right)
\end{aligned}
\label{chamfer_loss}
\end{equation}
where $x_i$ is the $i$-th point in the point cloud $x$, and
\begin{equation}
d_{\text{Chamfer}}(A,B)=\sum_{a\in A}\text{min}_{b\in B}\|a-b\|_2^2 + \sum_{b\in B}\text{min}_{a\in A}\|b-a\|_2^2
\end{equation}
is the symmetric Chamfer distance between $A,B\subset \mathbb{R}^3$. Since the Chamfer distance is computed only on within-region points, it does not burden the computation. 

In our experiments, we found that applying DefRec only to target samples yields better results. Therefore, unless stated otherwise, that is the selected approach.


\subsection{Point cloud mixup}\label{s:pcm}
We now discuss an additional contribution that is independent of the proposed SSL task, but we find to operate well with DefRec. The labeled samples from the source domain are commonly used in domain adaptation with a standard cross-entropy classification loss. Here, we suggest an alternative loss motivated by the Mixup Method \cite{zhang2018mixup}. Mixup is based on the Vicinal Risk Minimization principle, as opposed to Empirical Risk Minimization, and can be viewed as an extension of data augmentation that involves both input samples and their labels. Given two images and their "one-hot" labels $(x,y),(x',y')$, the Mixup method generates a new labeled sample as a convex combination of the inputs $(\gamma x + (1-\gamma)x',\gamma y + (1-\gamma)y')$, where $\gamma$ is sampled from a $Beta$ distribution with fixed parameters.

We generalize this method to point clouds. Since point clouds are arbitrarily ordered, a naive convex combination of two points (similarly to pixels) may result in arbitrary positions. Hence, such combination of two point clouds may be meaningless.
Instead, we propose the following \textit{Point Cloud Mixup} (PCM) procedure. Given two point clouds $x, x'\in \mathbb{R} ^{n\times 3}$, first sample a Mixup coefficient $\gamma \backsim Beta (\alpha, \beta)$ ($\alpha,\beta = 1$ worked well in our case) and then form a new shape by randomly sampling $\gamma \cdot n$ points from $x$ and $ (1-\gamma) \cdot n$ points from $x'$. The union of the sampled points yields a new point cloud, $\overline{x}\in \mathbb{R} ^{n\times 3}$. As in the original Mixup method, the label is a convex combination of the one-hot label vectors of the two point clouds $\gamma y + (1-\gamma) y'$. See \figref{fig:arch} (green box) for illustration.
 
To summarize, using PCM, the supervised data flow starts with sampling two labeled point clouds $(x,y),(x',y')\in S$. They are then combined into a new labeled point cloud $(\overline{x},\overline{y})$. 
$\overline{x}$ is fed into the shared encoder $\Phi$ to produce a point-cloud representation $\Phi(\overline{x})\in \mathbb{R}^d$. This representation is further processed by a fully connected sub-network (head) $h_{\text{sup}}$. A cross-entropy loss $L_{\text{ce}}$ is then applied to the output of $h_{\text{sup}}$ and the new label $\overline{y}$.

\begin{figure}[!t]
\centering
\includegraphics[width=0.35\textwidth]{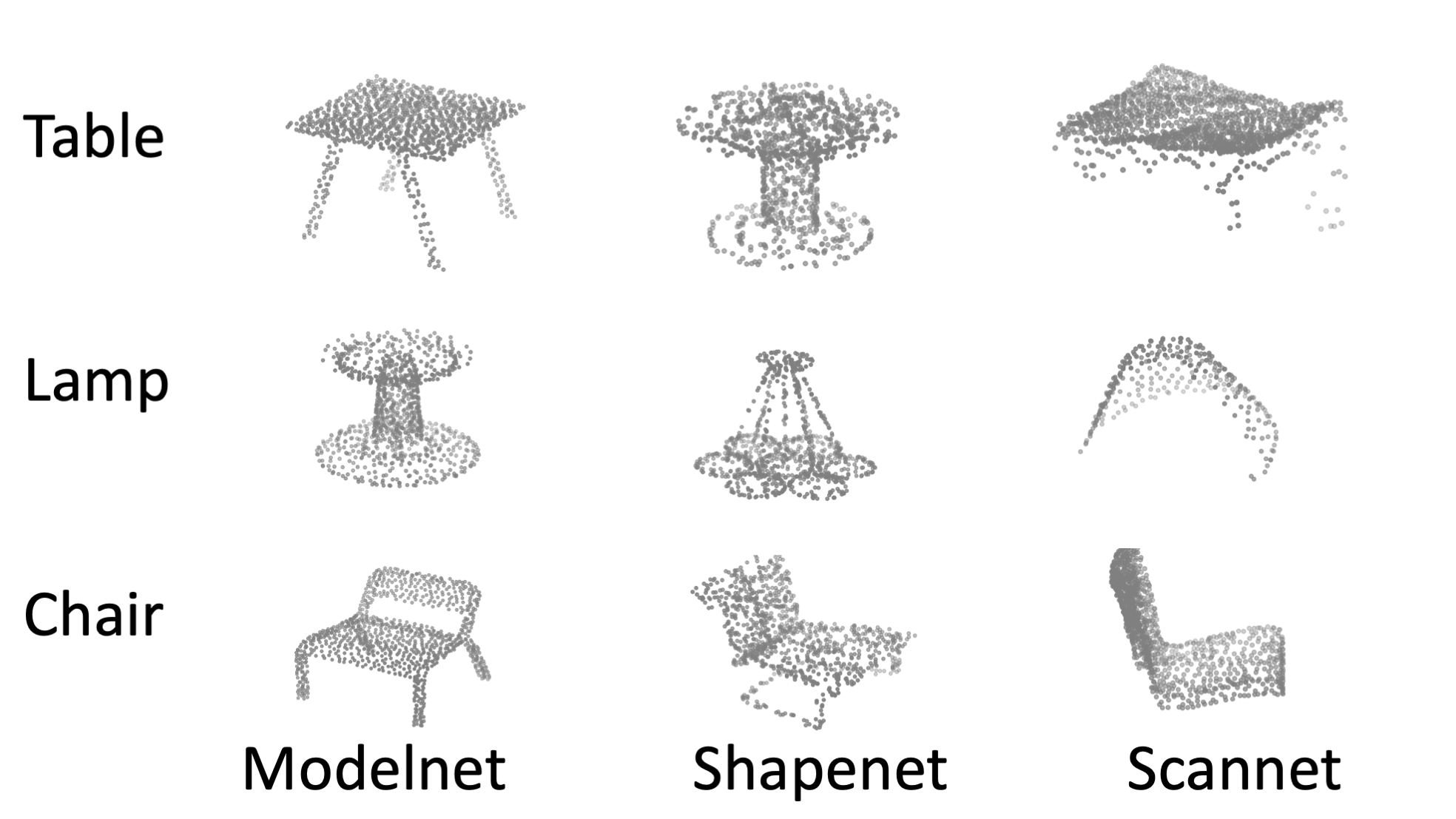}
\caption{A comparison of typical shapes from the datasets: ModelNet-10, ShapeNet-10, and ScanNet-10.}
\label{fig:datasets}
\end{figure}

\textbf{Mixup for DA.} Extensions of the Mixup method were offered as a solution for DA on images data \cite{mao2019virtual, xu2019adversarial, yan2020improve}. We, on the other hand, propose to use SSL methods, and in particular DefRec. Our formulation of Mixup for point-cloud data can be applied to any classification task and not necessarily for DA. We found that PCM improves the accuracy of various baselines ( \secref{sec:baselines_with_pcm} in the Appendix), and was particularly beneficial when combined with DefRec. 

\subsection{Overall loss}
The overall loss is a linear combination of a supervised loss and an SSL loss:
\begin{equation}
    \begin{aligned}
    L(S,T;\Phi,h_{\text{sup}},h_{\text{sup}}) &= \\
    & \hspace{-2.0cm} L_{\text{ce}}(S;\Phi,h_{\text{sup}}) +\lambda L_{\text{SSL}}(\widehat{S\cup T};\Phi,h_{\text{SSL}}),
    \end{aligned}
\end{equation}
where $\lambda$ is a parameter that controls the importance of the self-supervised term. To use PCM, $L_{\text{ce}}(S;\Phi,h_{\text{sup}})$ can be replaced with $L_{\text{ce}}(\overline{S};\Phi,h_{\text{sup}})$.

\begin{figure}[!t]
    \centering
    \includegraphics[width=0.3\textwidth]{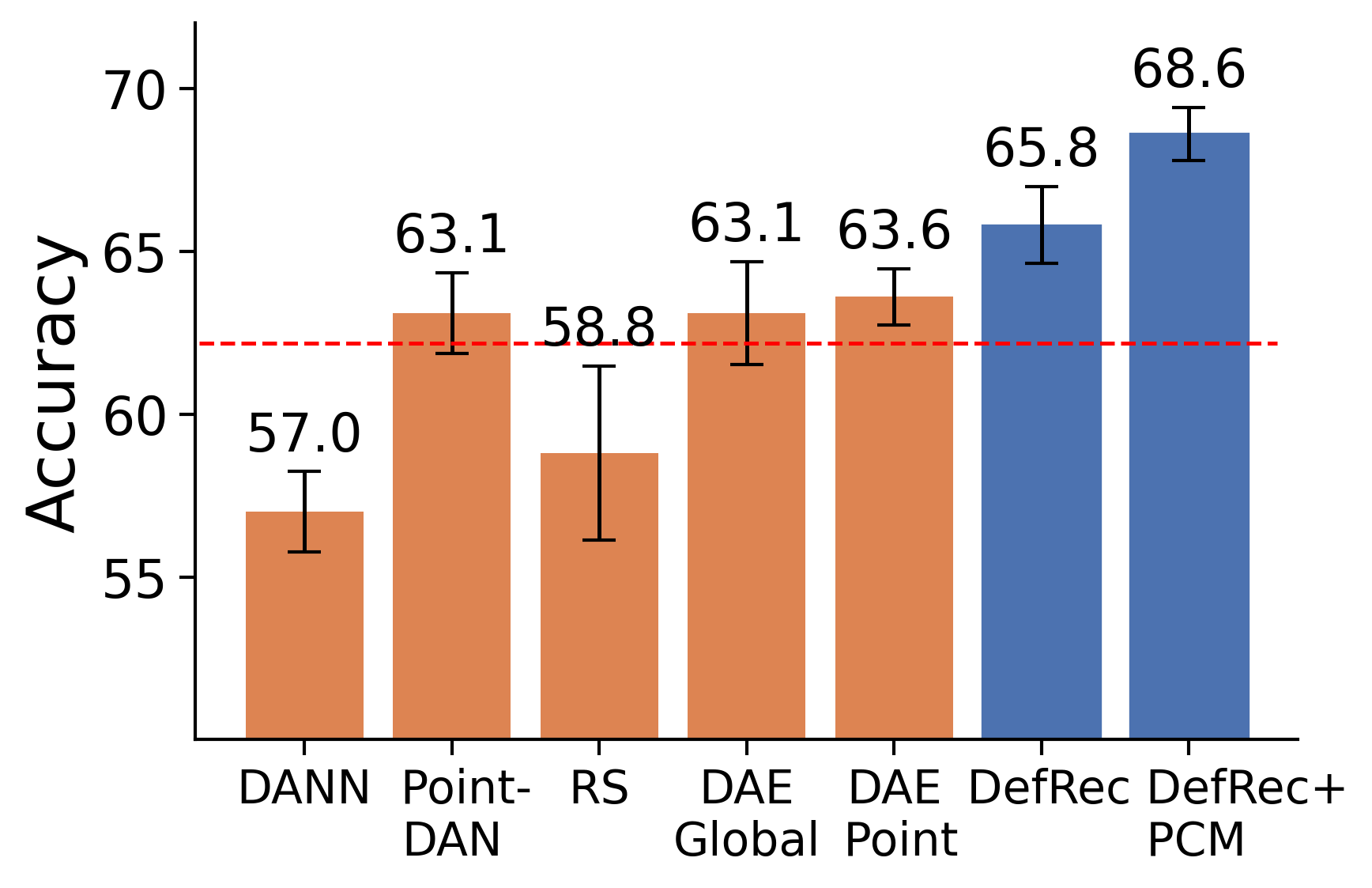}
    \caption{Average test accuracy on PointDA-10 dataset across six adaptations tasks. Dashed red line indicates the average accuracy of the unsupervised baseline.}
    \label{fig:avg_acc}
\end{figure}

\begin{table*}[!t]
  \centering
    \scalebox{0.8}{
    \begin{tabular}{p{10em}|p{6em}|p{6em}|p{6em}|p{6em}|p{6em}|p{6em}}
    \toprule
    \textbf{Method} & \textbf{ModelNet to \newline{}ShapeNet} & \textbf{ModelNet to \newline{}ScanNet} & \textbf{ShapeNet to \newline{}ModelNet} & \textbf{ShapeNet to \newline{}ScanNet} & \textbf{ScanNet to\newline{}ModelNet} & \textbf{ScanNet to \newline{}ShapeNet} \\
    \midrule
    \midrule
    Supervised-T & 93.9 $\pm$ 0.2 & 78.4 $\pm$ 0.6 & 96.2 $\pm$ 0.1 & 78.4 $\pm$ 0.6 & 96.2 $\pm$ 0.1 & 93.9 $\pm$ 0.2 \\
    Supervised & 89.2 $\pm$ 0.6 & 76.2 $\pm$ 0.6 & 93.4 $\pm$ 0.6 & 74.7 $\pm$ 0.7 & 93.2 $\pm$ 0.3 & 88.1 $\pm$ 0.7 \\
    \midrule
    Unsupervised & 83.3 $\pm$ 0.7 & 43.8 $\pm$ 2.3 & 75.5 $\pm$ 1.8 & 42.5 $\pm$ 1.4 & 63.8 $\pm$ 3.9 & 64.2 $\pm$ 0.8 \\
    DANN \cite{ganin2016domain}  & 75.3 $\pm$ 0.6 & 41.5 $\pm$ 0.2 & 62.5 $\pm$ 1.4 & 46.1 $\pm$ 2.8 & 53.3 $\pm$ 1.2 & 63.2 $\pm$ 1.2 \\
    PointDAN \cite{qin2019pointdan} & 82.5 $\pm$ 0.8 & 47.7 $\pm$ 1.0 & 77.0 $\pm$ 0.3 & 48.5 $\pm$ 2.1 & 55.6 $\pm$ 0.6 & 67.2 $\pm$ 2.7 \\
    RS \cite{sauder2019self}  & 81.5 $\pm$ 1.2 & 35.2 $\pm$ 5.9 & 71.9 $\pm$ 1.4 & 39.8 $\pm$ 0.7 & 61.0 $\pm$ 3.3 & 63.6 $\pm$ 3.4 \\
    DAE-Global \cite{hassani2019unsupervised} & \textbf{83.5 $\pm$ 0.8} & 42.6 $\pm$ 1.4 & 74.8 $\pm$ 0.8 & 45.5 $\pm$ 1.6 & 64.9 $\pm$ 4.4 & 67.3 $\pm$ 0.6 \\
    DAE-Point & 82.5 $\pm$ 0.4 & 40.2 $\pm$ 1.6 & 76.4 $\pm$ 0.7 & 50.2 $\pm$ 0.5 & 66.3 $\pm$ 1.5 & 66.1 $\pm$ 0.5 \\
    \midrule
    DefRec (ours) & 83.3 $\pm$ 0.2  & 46.6 $\pm$ 2.0  &\textbf{79.8 $\pm$ 0.5} & 49.9 $\pm$ 1.8 & 70.7 $\pm$ 1.4 & 64.4 $\pm$ 1.2 \\
    DefRec + PCM (Ours) & 81.7 $\pm$ 0.6 & \textbf{51.8 $\pm$ 0.3} & 78.6 $\pm$ 0.7 & \textbf{54.5 $\pm$ 0.3} & \textbf{73.7 $\pm$ 1.6} & \textbf{71.1 $\pm$ 1.4} \\
   \bottomrule
    \end{tabular}%
    }
  \caption{Test accuracy on PointDA-10 dataset, averaged over three runs ($\pm$ SEM).}
  \label{table:main_table}%
\end{table*}%

\section{Experiments}\label{section:exp}
We evaluate our method on \textit{PointDA-10}, a DA dataset designed by \cite{qin2019pointdan} for classification of point clouds. For segmentation, we introduce a benchmark dataset, \textit{PointSegDA}. 

\subsection{PointDA-10}
PointDA-10 (\cite{qin2019pointdan}) consists of three subsets of three widely-used datasets: ShapeNet \cite{chang2015shapenet}, ModelNet \cite{wu20153d} and ScanNet \cite{dai2017scannet}. All three subsets share the same ten distinct classes (like chair, table, bed). 
\textit{ModelNet-10} (called ModelNet hereafter) contains 4183 train samples and 856 test samples sampled from clean 3D CAD models. \textit{ShapeNet-10} (called ShapeNet hereafter), contains 17,378 train samples and 2492 test samples sampled from several online repositories of 3D CAD models. 
\textit{ScanNet-10} (called ScanNet hereafter) contains 6110 train and 1769 test samples. ScanNet is an RGB-D video dataset of scanned real-world indoor scenes. 
Samples from this dataset are significantly harder to classify because: (i) Many objects have missing parts, due to occlusions and, (ii) some objects are sampled sparsely. See \figref{fig:datasets} for a comparison of typical shapes from all the datasets mentioned above. Further details about the experimental setup are provided in \secref{sec:imp_details} in the Appendix.

\subsection{PointSegDA}
For evaluating point cloud segmentation we built a new benchmark dataset called PointSegDA. It is based on a dataset of meshes of human models proposed by \cite{maron2017convolutional}, which consists of four subsets: ADOBE, FAUST, MIT, and SCAPE. They share eight classes of human body parts (feet, hand, head, etc.) but differ in point distribution, pose and, scanned humans. Point clouds were extracted and labeled from the meshes to accommodate our setup. 
PointSegDA differs from PointDA-10 in the type of the domain shifts (different humans, poses and discretizations in each subset), the actual shapes and in the fact that it represent deformable objects. Thus, PointSegDA allows us to show the applicability of DefRec in a fundamentally different setup. Further details are provided in Appendix \ref{sec:PointSegDA_dataset} and \ref{sec:pointSegda_imp}. 

\subsection{Network architecture}
The input to the network is a point cloud with 1024 points (for PointDA-10) or 2048 points (for PointSegDA). For a feature extractor and the supervised task head, we used DGCNN \cite{wang2019dynamic} with the same configurations as in the official implementation. As for the SSL head, differently from common solutions in the literature (\eg \cite{achlioptas2018learning, hassani2019unsupervised}), it takes as input the global feature vector (of size 1024) concatenated to the feature representations of the points from the backbone network. We consistently found that it generates better solutions. Additional details on the architecture are presented in Appendix \ref{sec:imp_details}.

\section{Results}\label{s:results}
We now discuss the results of using DefRec and PCM. The same pre-processing and experiemntal setup was applied to all methods (ours and baselines). In all experiments we report the mean accuracy and standard error of the mean (SEM) across three runs with different seeds. For each of the three deformation types of DefRec we examined different hyper-parameters, such as radii size for volume-based deformations or layer depth for feature-based deformations. See detailed explanation in Appendix \ref{sec:imp_details}.

\begin{table*}[!t]
  \centering
  \scalebox{0.8}{
    \begin{tabular}{p{6em}|p{6em}|p{6em}|p{6em}|p{6em}|p{6em}|p{6em}|p{5em}}
    \toprule
    \textbf{Deformation Type} & \textbf{ModelNet to \newline{}ShapeNet} & \textbf{ModelNet to \newline{}ScanNet} & \textbf{ShapeNet to \newline{}ModelNet} & \textbf{ShapeNet to \newline{}ScanNet} & \textbf{ScanNet to\newline{}ModelNet} & \textbf{ScanNet to \newline{}ShapeNet} & \textbf{Avg.}\\
    \midrule
    Volume-based & 81.7 $\pm$ 0.6& 51.8 $\pm$ 0.3 & \textbf{78.6 $\pm$ 0.7} & \textbf{54.5 $\pm$ 0.3} & 73.7 $\pm$ 1.6 & 71.1 $\pm$ 1.4 & 68.6 $\pm$ 0.8\\
    \midrule
    Feature-based & 83.8 $\pm$ 0.8 & 44.3 $\pm$ 0.7 & 75.6 $\pm$ 1.0 & 52.2 $\pm$ 0.7 & \textbf{74.0 $\pm$ 1.7} & \textbf{77.2 $\pm$ 0.5} & 67.9 $\pm$ 0.9 \\
    \midrule
    Sample-based & \textbf{85.0 $\pm$ 0.5} & 44.6 $\pm$ 2.0 & 72.3 $\pm$ 1.9 & 52.1 $\pm$ 0.1 & 73.3 $\pm$ 0.7 & 74.3 $\pm$ 0.7 & 66.9 $\pm$ 1.0\\
    \bottomrule
    \end{tabular}%
    }
  \caption{Performance per deformation type. Test accuracy on PointDA-10 dataset, averaged over three runs ($\pm$ SEM).}
  \label{table:accuracy_by_family}%
\end{table*}%

\subsection{Classification accuracy}
We compared DefRec with the following baselines: (1) \textit{Unsupervised}, using only labeled source samples. (2) \textit{DANN} \cite{ganin2016domain}, a baseline commonly used in DA for images. Training with a domain classifier to distinguish between source and target. (3) \textit{PointDAN} \cite{qin2019pointdan} that suggested to align features both locally and globally. The global feature alignment is implemented using the method proposed in \cite{saito2018maximum} and therefore this baseline can also be seen as an extension of it. (4) \textit{RS}, using the SSL task for point clouds suggested in \cite{sauder2019self} instead of DefRec. Split the space to $3 \times 3 \times 3$ equally sized voxels, shuffling them and, assign the network to reconstruct the original order. (5) \textit{Denoising Auto-Encoder (DAE)-Global} \cite{hassani2019unsupervised}, reconstruction from a point cloud perturbed with i.i.d Gaussian noise. Since this method proposed to reconstruct from a global feature vector we also compared to (6) \textit{DAE-Point}, reconstruction from a point cloud perturbed with i.i.d. Gaussian noise with the same input to $h_{\text{SSL}}$ as DefRec; a concatenation of the global feature vector to the point features. We also present two upper bounds: (1) \textit{Supervised-T}, training on target domain only and, (2) \textit{Supervised}, training with labeled source and target samples. DefRec, RS, DAE-Point and DANN all have 2.5M parameters, compared with 11.1M in PointDAN and 12.6M parameters in DAE-Global.

We test several deformation types, each with its own variants (such as the radius size for volume-based deformations). To get a unified measure of performance, we treated the deformation type and its variants as hyperparameters. Then, for each adaptation, we followed a stringent protocol and picked the model that maximized accuracy on the \textit{validation set of the source} distribution. 

Table~\ref{table:main_table} shows the classification accuracy of all methods on the six adaptation tasks. Figure~\ref{fig:avg_acc} shows the average accuracy per method across the six adaptation tasks. As can be seen, our methods outperform all baselines in 5 out of 6 adaptations. The average across six adaptations of both of our methods (DefRec and DefRec + PCM) is the highest. DefRec + PCM improve by 5\% compared to the best baseline and by 5.5\% compared to PointDAN, the natural competitor on this dataset. Also, DefRec is more accurate on sim-to-real adaptations (\textit{ModelNet-to-ScanNet} and \textit{ShapeNet-to-Scannet}). This observation validates our intuition: (i) DefRec promotes learning semantic properties of the shapes and, (ii) DefRec helps the model in generalizing to real data that has missing regions/parts.

Appendix~\ref{sec:baselines_with_pcm} further quantifies the effect of combining PCM with baseline methods. PCM boosts the performance of several baselines. However, our proposed DefRec+PCM is still superior. Appendix~\ref{sec:mixing_def} present a variant of DefRec in which the deformation types are combined efficiently.

\begin{table*}[t]
  \centering
  \scalebox{0.8}{
    \begin{tabular}{p{10em}|p{6em}|p{6em}|p{6em}|p{6em}|p{6em}|p{6em}|p{5em}}
    \toprule
    \textbf{Method} & \textbf{ModelNet to \newline{}ShapeNet} & \textbf{ModelNet to \newline{}ScanNet} & \textbf{ShapeNet to \newline{}ModelNet} & \textbf{ShapeNet to \newline{}ScanNet} & \textbf{ScanNet to\newline{}ModelNet} & \textbf{ScanNet to \newline{}ShapeNet} & \textbf{Avg.}\\
    \midrule
    PCM only & \textbf{83.7 $\pm$ 0.6} & 42.6 $\pm$ 0.9 & 71.4 $\pm$ 1.5 & 46.1 $\pm$ 1.7 & 71.5 $\pm$ 1.0 & 74.6 $\pm$ 0.5 & 65.0 $\pm$ 1.0\\
    \midrule
    DefRec only & 82.7 $\pm$ 0.6 & 43.9 $\pm$ 1.3 &\textbf{79.8 $\pm$ 0.5} & 48.0 $\pm$ 0.6 & 66.0 $\pm$ 0.8 & 67.4 $\pm$ 1.2 & 64.6 $\pm$ 0.8\\
    \midrule
    DefRec Global + PCM & 82.1 $\pm$ 0.5 & 50.1$\pm$ 3.1 & 75.0 $\pm$ 1.3 & 51.6 $\pm$ 1.6 & 61.1 $\pm$ 4.4 & \textbf{76.3 $\pm$ 1.0} & 66.0 $\pm$ 2.0\\
    \midrule
    DefRec S/T + PCM & 82.6 $\pm$ 0.6 & 53.1 $\pm$ 1.0 & 78.3 $\pm$ 1.0 & 51.5 $\pm$ 0.9 & 72.0 $\pm$ 0.5 & 74.4 $\pm$ 0.8 & 68.7 $\pm$ 0.8\\
    \midrule
    DefRec + PCM & 83.3 $\pm$ 0.1 & \textbf{53.5 $\pm$ 1.6} & 78.5 $\pm$ 0.6 & \textbf{53.2 $\pm$ 1.8} & \textbf{73.7 $\pm$ 0.6} & 75.5 $\pm$ 0.9 & \textbf{69.6 $\pm$ 0.9}\\
    \bottomrule
    \end{tabular}%
    }
  \caption{Ablation study \& model configurations. Test accuracy on PointDA-10 dataset, averaged over three runs ($\pm$ SEM).} 
  \label{abblation}%
\end{table*}%

\subsection{Analysis}
We now analyze the three deformation types of DefRec. See Appendix \ref{sec:perplexity} and \ref{sec:shape_rec_supp} for further analysis of the representation learned and demonstration of shapes reconstruction.

\subsubsection{Accuracy by deformation category}
Table~\ref{table:accuracy_by_family} shows the test accuracy for the three types of deformations: Volume-based, Feature-based and Sample-based. For each type, per adaptation, we selected the best model among all variants of that family according to source-validation accuracy. As seen from the table, deforming based on proximity in the input space yields the highest accuracy on average. In fact, across all adaptations, variants of the volume-based deformation type also had the highest source-validation accuracy. Note that when considering the three types of deformation separately, namely considering each type of deformation as a separate method (unlike the results in Table~\ref{table:main_table} in which we treated the type as a hyper-parameter), DefRec has the highest accuracy across all adaptation tasks. 

\begin{table*}[!t]
    \centering 
    \scalebox{0.64}{
    \setlength{\tabcolsep}{1pt} 
    \hskip-0.2cm
    \begin{tabular} 
    {p{9em}|p{5em}|p{5em}|p{5em}|p{5em}|p{5em}|p{5em}|p{5em}|p{5em}|p{5em}|p{5em}|p{5em}|p{5em}|p{5em}}
    \toprule
    \textbf{Method} & \textbf{FAUST to \newline{}ADOBE} & \textbf{FAUST to \newline{} MIT} & \textbf{FAUST to \newline{}SCAPE} & \textbf{MIT to \newline{}ADOBE} & \textbf{MIT to \newline{} FAUST} & \textbf{MIT to \newline{}SCAPE} & \textbf{ADOBE to \newline{}FAUST} & \textbf{ADOBE to \newline{} MIT} & \textbf{ADOBE to \newline{}SCAPE} & \textbf{SCAPE to \newline{}ADOBE} & \textbf{SCAPE to \newline{} FAUST} & \textbf{SCAPE to \newline{}MIT} & \textbf{Avg.}\\
    \midrule
    \midrule
    Supervised-T & 80.9 $\pm$ 7.2 & 81.8 $\pm$ 0.3 & 82.4 $\pm$ 1.2 & 80.9 $\pm$ 7.2 & 84.0 $\pm$ 1.8 & 82.4 $\pm$ 1.2 & 84.0 $\pm$ 1.8 & 81.8 $\pm$ 0.3 & 82.4 $\pm$ 1.2 & 80.9 $\pm$ 7.2 & 84.0 $\pm$ 1.8 & 81.8 $\pm$ 0.3 & 82.3 $\pm$ 2.6\\
    \midrule
    Unsupervised & 78.5 $\pm$ 0.4 & 60.9 $\pm$ 0.6 & 66.5 $\pm$ 0.6 & 26.6 $\pm$ 3.5 & 33.6 $\pm$ 1.3 & 69.9 $\pm$ 1.2 & 38.5 $\pm$ 2.2 & 31.2 $\pm$ 1.4 & 30.0 $\pm$ 3.6 & \textbf{74.1 $\pm$ 1.0} & 68.4 $\pm$ 2.4 & 64.5 $\pm$ 0.5 & 53.6 $\pm$ 1.6\\
    Adapt-SegMap \cite{tsai2018learning} & 70.5 $\pm$ 3.4 & 60.1 $\pm$ 0.6 & 65.3 $\pm$ 1.3 & 49.1 $\pm$ 9.7 & \textbf{54.0 $\pm$ 0.5} & 62.8 $\pm$ 7.6 & 44.2 $\pm$ 1.7 & \textbf{35.4 $\pm$ 0.3} & 35.1 $\pm$ 1.4 & 70.1 $\pm$ 2.5 & 67.7 $\pm$ 1.4 & 63.8 $\pm$ 1.2 & 56.5 $\pm$ 2.6\\
    RS \cite{sauder2019self} & 78.7 $\pm$ 0.5 & 60.7 $\pm$ 0.4 & 66.9 $\pm$ 0.4 & 59.6 $\pm$ 5.0 & 38.4 $\pm$ 2.1 & 70.4 $\pm$ 1.0 & 44.0 $\pm$ 0.6 & 30.4 $\pm$ 0.5 & 36.6 $\pm$ 0.8 & 70.7 $\pm$ 0.8 & \textbf{73.0 $\pm$ 1.5} & 65.3 $\pm$ 0.1 & 57.9 $\pm$ 1.1\\
    \midrule
    DefRec (ours) & \textbf{79.7 $\pm$ 0.3} & \textbf{61.8 $\pm$ 0.1} & \textbf{67.4 $\pm$ 1.0} & \textbf{67.1 $\pm$ 1.0} & 40.1 $\pm$ 1.4 & \textbf{72.6 $\pm$ 0.5} & 42.5 $\pm$ 0.3 & 28.9 $\pm$ 1.5 & 32.2 $\pm$ 1.2 & 66.4 $\pm$ 0.9 & 72.2 $\pm$ 1.2 & \textbf{66.2 $\pm$ 0.9} & \textbf{58.1 $\pm$ 0.9}\\
    DefRec + PCM (Ours) & 78.8 $\pm$ 0.2 & 60.9 $\pm$ 0.8 & 63.6 $\pm$ 0.1 & 48.1 $\pm$ 0.4 & 48.6 $\pm$ 2.4 & 70.1 $\pm$ 0.8 & \textbf{46.9 $\pm$ 1.0} & 33.2 $\pm$ 0.3 & \textbf{37.6 $\pm$ 0.1} & 66.3 $\pm$ 1.7 & 66.5 $\pm$ 1.0 & 62.6 $\pm$ 0.2 & 56.9 $\pm$ 0.7\\
   \bottomrule
    \end{tabular}%
    }
  \caption{Test mean IoU on PointSegDA dataset, averaged over three runs ($\pm$ SEM).}
  \label{table:PointSegDA}%
\end{table*}%

\subsubsection{Volume-based deformations} \label{s:part_analysis}
The effectiveness of volume-based deformation is sensitive to the size of a deformed region. Small regions may be too easy for the network to reconstruct, while large regions may be very hard. To quantify this sensitivity  \figref{fig:acc_diff_by_radius} shows the mean accuracy gain averaged over 6 adaptations as a function of deformation radius. All values denote the gain compared with a baseline of deforming the full shape ($r\!\!=\!\!2.0$). Accuracy is maximized with small to mid-sized regions, with an optimum at $r\!=\!0.2$. This suggest that deformations at the scale of object parts are superior to global deformations, in agreement with the intuition that mid-scale regions capture the semantic structures of objects.

\subsubsection{Feature-based deformations} \label{s:feature_analysis}
\figref{fig:layer_num_pts_effect} traces the accuracy as a function of the number of points when deforming based on proximity in feature space. As in \secref{s:part_analysis}, we find that deforming large regions degrades the performance, particularly with more than 300 points. Also, layer 4 is dominated by layer 3 by a small gap. Overall, the model is largely robust to the choice of layer and the number of points for small enough regions.


\subsubsection{SSL vs data augmentation} \label{s:sample_analysis}
In this paper we promote the use SSL tasks for bridging the domain gap. An interesting question arises: could this gap be bridged using deformations as a data augmentation mechanism? In this case, we may use only the labeled source samples for supervision. As an example, consider a sim-to-real adaptation task. The sampling procedures suggested in this paper can be used for data augmentation. These methods sample some parts of the object more densely and other parts more sparsely. As a result, the augmented shapes may resemble to shapes from the target data.

\begin{figure}[!t]
    \centering
    \subfloat[[Volume-based def.][Volume-based def.]{
    \includegraphics[width=0.32\linewidth]{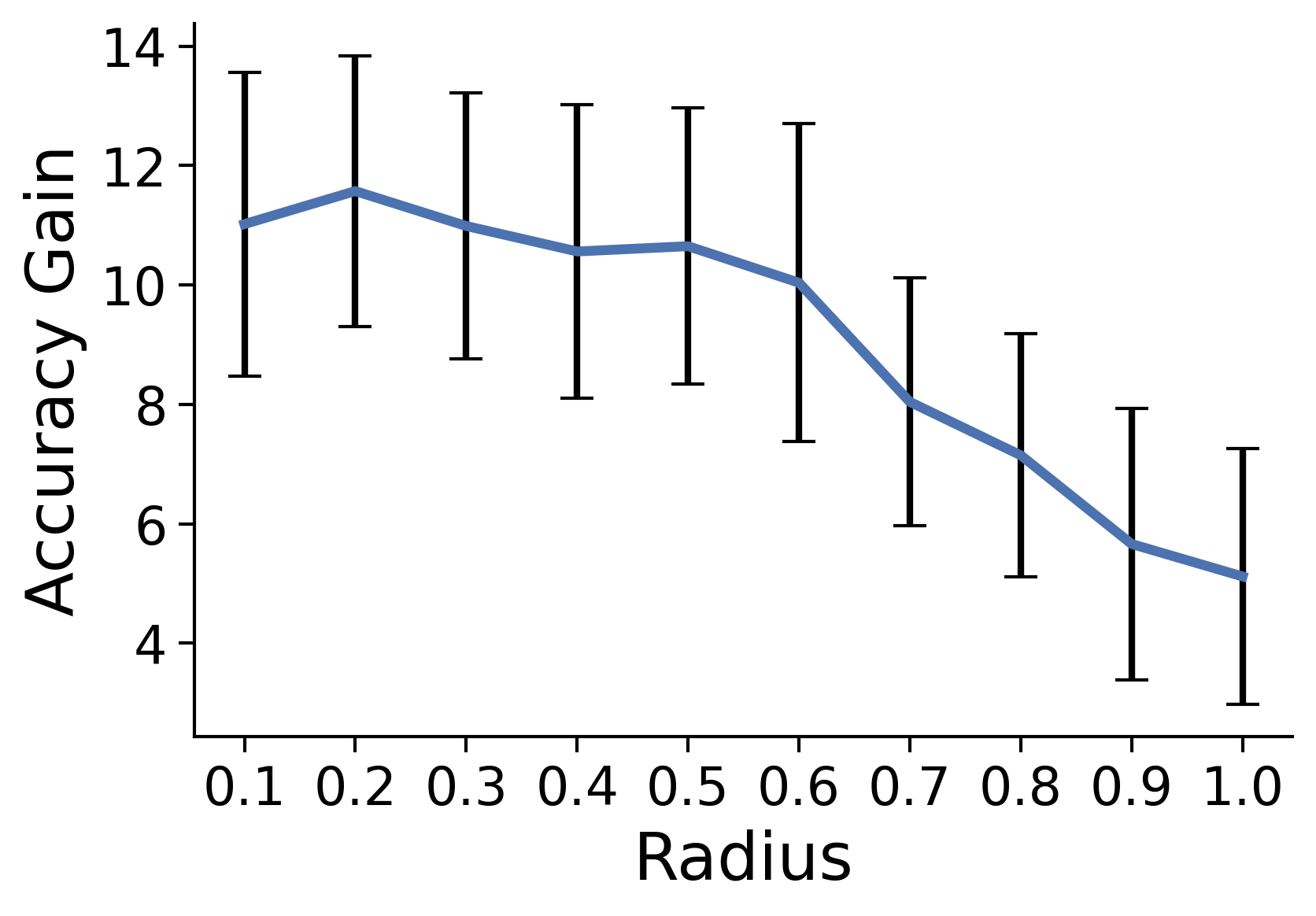}
    \label{fig:acc_diff_by_radius}}
    \subfloat[Feature-based def.]{
    \includegraphics[width=0.32\linewidth]{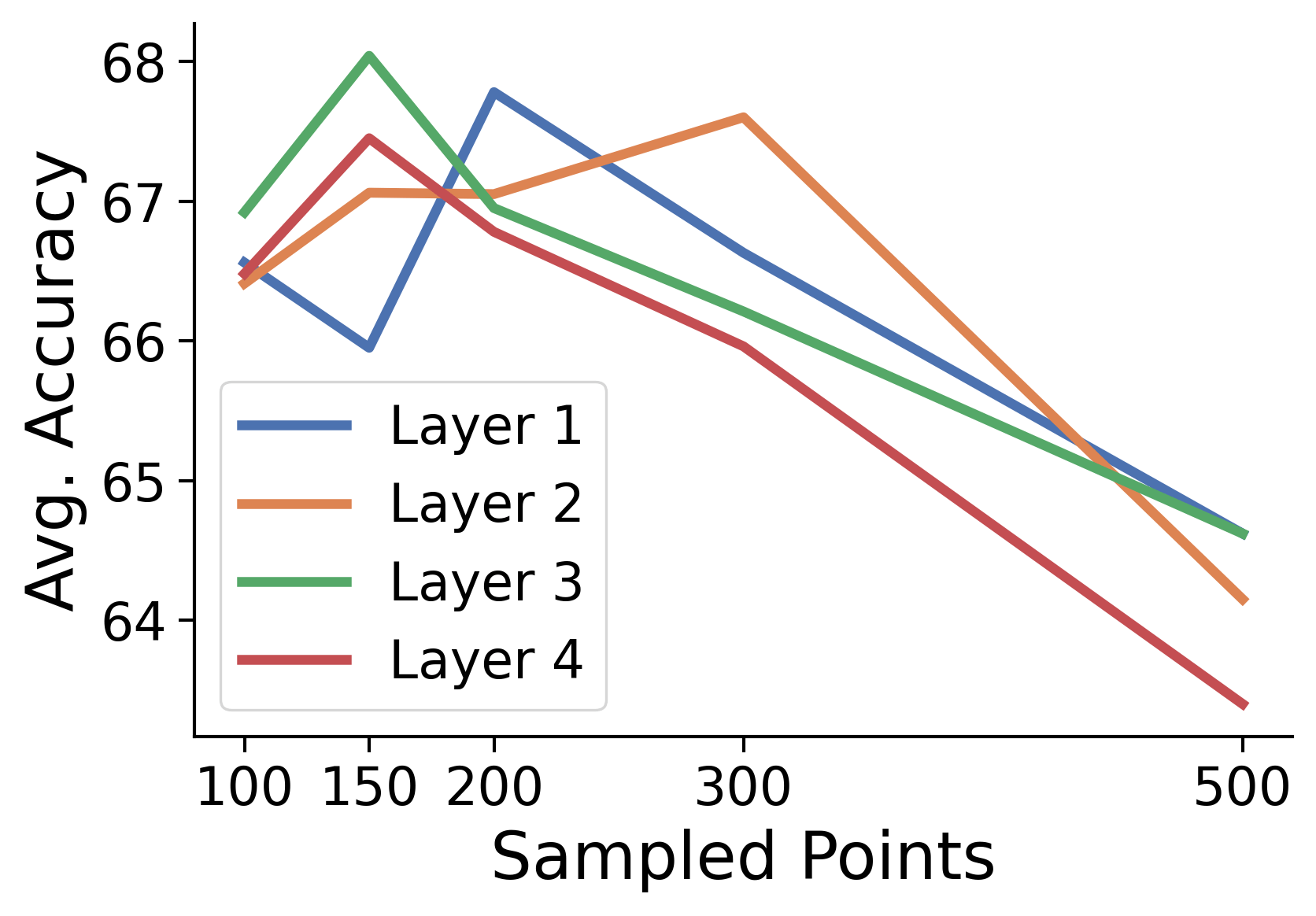}
    \label{fig:layer_num_pts_effect}}
    \subfloat[Sample-based def.][Sample-based def.]{
    \includegraphics[width=0.32\linewidth]{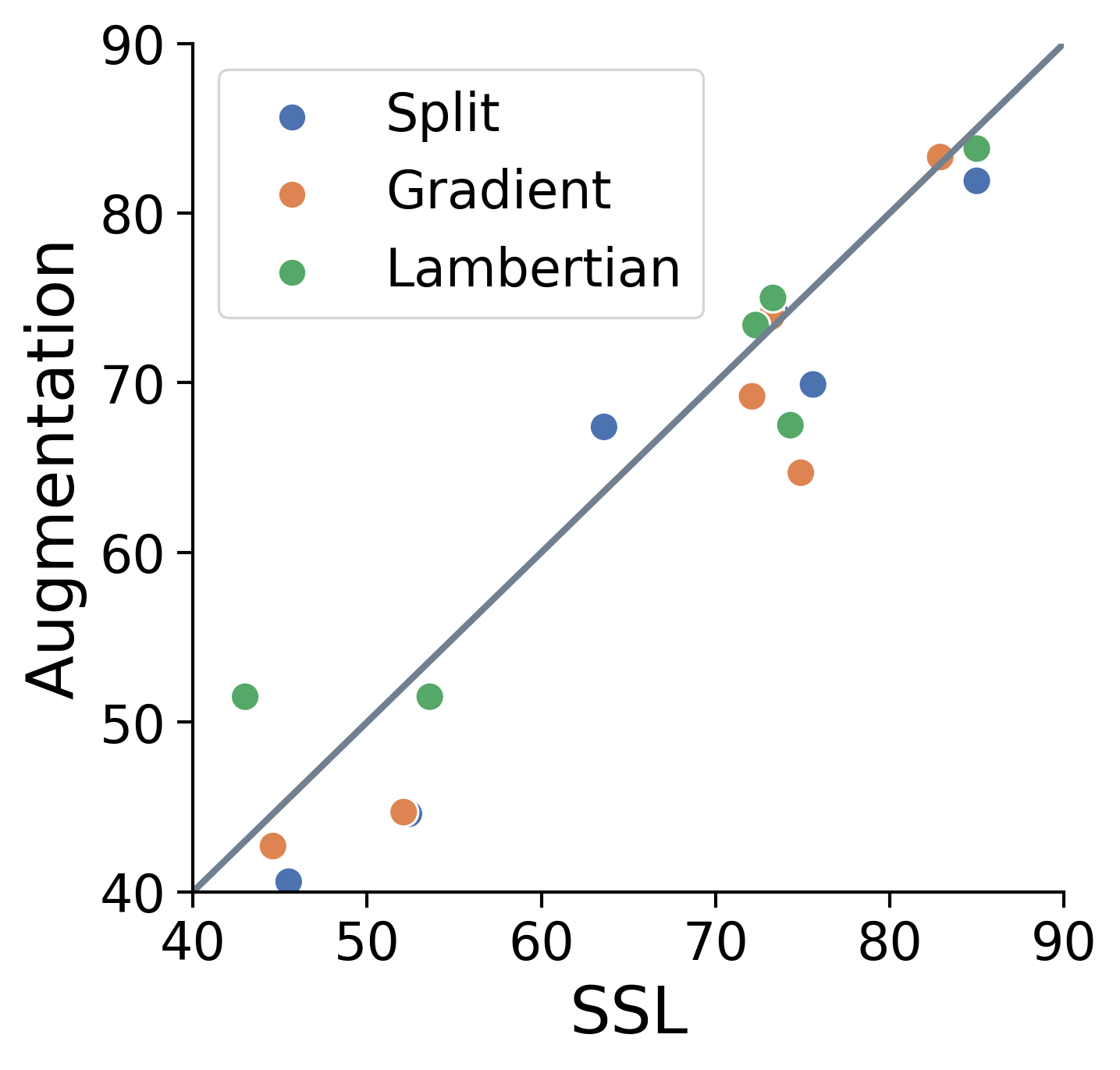}
    \label{fig:ssl_vs_aug}}

    \caption{Analysis of the deformation approaches. \textbf{(a)} Classification accuracy as a function of the deformation radius. 
    Shown is the average ($\pm$ SEM) gain in accuracy compared with the accuracy for radius 2.0.
    \textbf{(b)} Classification accuracy as a function of the deformation size and shared feature encoder layer. Each curve corresponds to the accuracy averaged across six adaptation tasks of a different layer from the shared feature encoder. \textbf{(c)} Self-supervised learning vs data augmentation. Each marker is the average accuracy (across three seeds) for a specific combination of adaptation (out of 6) and deformation (out of 3). Total of 18 points.
    }
    \label{fig:Analysis_families}
\end{figure}

To test this idea we use the sample-based deformations in two fashions: (i) As an SSL task, the method advocated in this paper and, (ii) as a data augmentation procedure for source samples. Figure~\ref{fig:ssl_vs_aug} compares these alternatives on the six adaptations tasks with the three sampling procedures. Most data points (11/18) are below the diagonal line $y=x$, five of which are on sim-to-real adaptations tasks. This result suggests that using the sampling procedures as an SSL task should be preferred over data augmentation.

\subsection{Ablation and additional experiments} \label{section:ablation}
To gain insight into the relative contribution of model components, we evaluate variants of our approach where we isolate the individual contribution of different components. We do so for the method in which we split the space to $3\times3\times3$ voxels from the volume-based type. 
Table~\ref{abblation} compares the following models: (1) \textit{DefRec-only}, applying DefRec to target data, no PCM. (2) \textit{PCM only}, applying PCM to source data, no DefRec. (3) \textit{DefRec Global + PCM} our method when reconstructing from a global feature vector, following~\cite{achlioptas2018learning, hassani2019unsupervised}. (4) \textit{DefRec-S/T + PCM} applying PCM to source data and DefRec to both source and target data and; (5) \textit{DefRec + PCM}, our proposed method of applying PCM to source data and DefRec to target data.

The results suggest: (a) When DefRec and PCM are considered independently, no module consistently outperforms the other, yet when using both modules there is a significant boost in most adaptation setups and in the overall performance. (b) Applying DefRec on both source and target samples degrades performance. (c) Performance often drops when reconstructing from the global feature vector.

\subsection{DefRec for segmentation} \label{sec:DefRec_for_seg}
To examine how DefRec generalizes beyond classification tasks, we tested it in segmentation tasks using PointSegDA dataset. It is easy to extend DefRec to this task. Similarly to classification, the network is trained jointly with two losses: one to segment labeled source objects and the second to reconstruct deformed target objects. Since labels are now provided per point, not per object, we adapted PCM to segmentation in the following way. Similarly to PCM for classification, we generate a new point cloud from a random split of two point clouds in the batch. Here, however, each point from the original shape migrates with its associated label. The overall loss is the mean cross-entropy calculated for all points in the new mixed shape. 

Table \ref{table:PointSegDA} compares the mean Intersection over Union (IoU) of DefRec with the unsupervised baseline, RS \cite{sauder2019self}, and Adapt-SegMap \cite{tsai2018learning}, a strong DA baseline for semantic segmentation of images which applies adversarial training in the output space.
The experiment shows that: (i) Adding auxiliary SSL tasks helps to bridge the domain gap for segmentation and, (ii) DefRec (either with or without PCM) achieves the highest results on most adaptations, validating the applicability of DefRec to segmentation tasks.

\section{Conclusions}
We tackled the problem of domain adaptation on 3D point clouds. 
We designed \textit{DefRec}, a novel family of self-supervised pretext tasks inspired by the kind of deformations encountered in real 3D point cloud data. In addition, we designed \textit{PCM}, a new training procedure for 3D point clouds based on the Mixup method that can be applied to any classification or segmentation task. PCM is complementary to DefRec, and when combined they form a strong model with relatively simple architecture. We demonstrated the benefit of our method on several adaptation setups, reaching a new state of the art results.

\subsection*{Acknowledgments} IA was funded by the Israel innovation authority as part of the AVATAR consortium, and by a grant from the Israel Science Foundation (ISF 737/2018). 

\clearpage
{\small
\bibliographystyle{ieee_fullname}
\bibliography{egbib}
}

\clearpage



\section*{Appendices}
\appendix
\section{PointSegDA dataset} \label{sec:PointSegDA_dataset}
\begin{figure}[!t]
    \centering
    \includegraphics[width=0.50\textwidth]{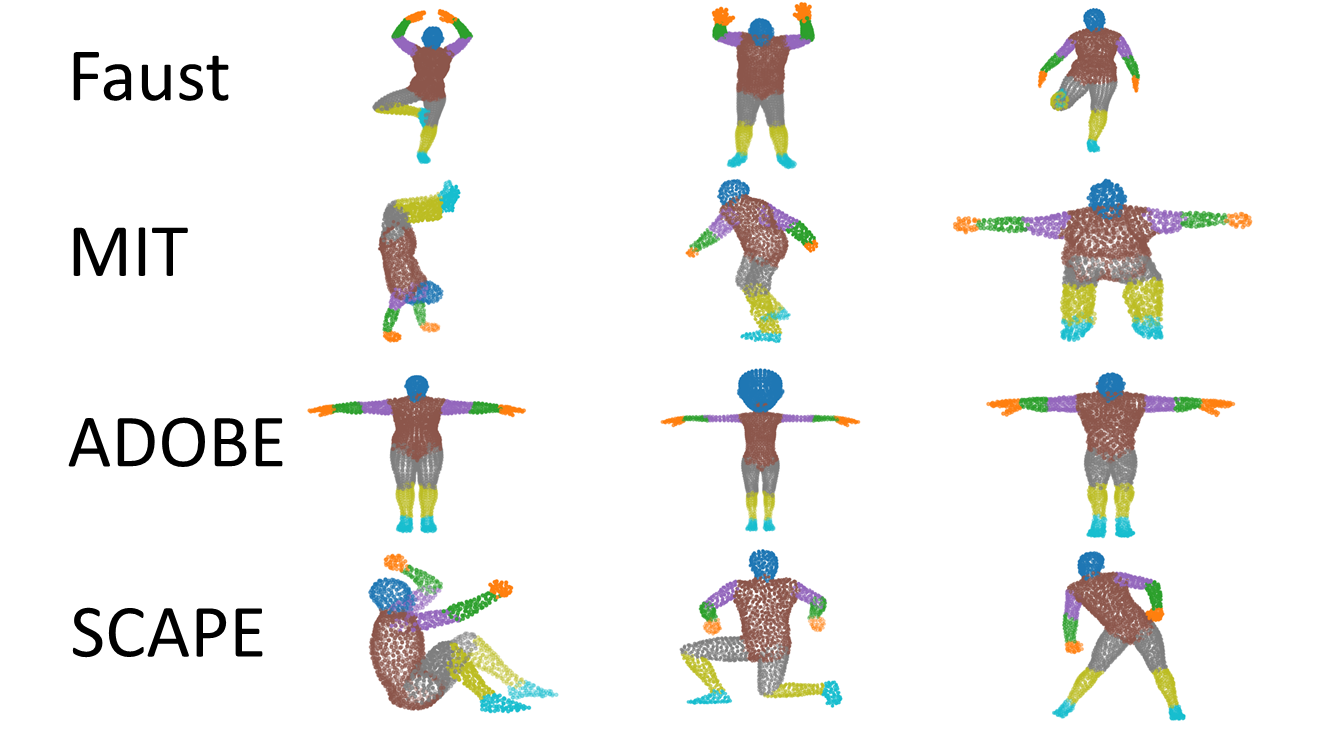}
    \caption{A comparison of typical shapes from the datasets: FAUST, MIT, ADOBE, and SCAPE.}
    \label{fig:PointSegDA_shapes}
\end{figure}

We built PointSegDA dataset based on a dataset of triangular meshes of human models proposed by \cite{maron2017convolutional}. This dataset is made of the following four datasets, which serve as different domains: ADOBE, FAUST, MIT, and SCAPE. The datasets differ in body pose, shape, and point distribution. We generated a point cloud from each mesh in each domain by extracting all the vertices (edges were neglected) and sampling 2048 points according to farthest point sampling. We aligned the shapes with the positive Z-axis and scaled them to the unit cube as in \cite{qi2017pointnet}.
Point labels were obtained from the polygon labels. In case of a conflict (i.e., the same point appears in two polygons having different labels) we randomly picked one label. As a result, a total of eight point-labels were obtained: foot, leg, thigh, torso, upper arm, forearm, hand, and head.
Table \ref{table:PointSegDA_split} shows the train/val/test splits, and Fig. \ref{fig:PointSegDA_shapes} depict three shapes from each domain.

\begin{table}[!h]
  \centering
  \scalebox{0.8}{
    \begin{tabular}{c|c|c|c|c}
    \toprule
    \textbf{Dataset} & \textbf{Train} & \textbf{Validation} & \textbf{Test} & \textbf{Total}\\
    \midrule
    FAUST & 70 & 10 & 20 & 100\\
    MIT & 118 & 17 & 34 & 169\\
    ADOBE & 29 & 4 & 8 & 42\\
    SCAPE & 50 & 7 & 14 & 71\\
    \bottomrule
    \end{tabular}%
    }
  \caption{Number of samples in each sets.}
  \label{table:PointSegDA_split}%
\end{table}%

\section{Implementation details} \label{sec:imp_details}
\subsection{PointDA-10 dataset} \label{sec:pointda_imp}

\textbf{Data processing.}
Following several studies \cite{qi2017pointnet,wang2019dynamic,li2018pointcnn} we assume that the upwards direction of all point clouds in all datasets is known and aligned. Since point clouds in ModelNet are aligned with the positive $Z$ axis, we aligned samples from ShapeNet and ScanNet in the same direction by rotating them about the x-axis. We sampled 1024 points from shapes in ModelNet and ScanNet (which have 2048 points) using farthest point sampling as in \cite{qi2017pointnet}. We split the training set to 80\% for training and 20\% for validation and scaled the shapes to the unit-cube. During training, we applied jittering as in \cite{qi2017pointnet} with standard deviation and clip parameters of 0.01 and 0.02 respectively, and random rotations to shapes about the $Z$ axis only.

\textbf{Network architecture.} 
In all methods we used DGCNN \cite{wang2019dynamic} for the feature extractor with the following configurations: Four point cloud convolution layers of sizes [64, 64, 128, 256] respectively and a 1D convolution layer with kernel size 1 (feature-wise fully connected) with a size of 1024 before extracting a global feature vector by max-pooling. 
We implemented a spatial transformation network to align the input point set to a canonical space using two point-cloud convolution layers with sizes [64, 128] respectively, a 1D convolution layer of size 1024 and three fully connected layers of sizes [512, 256, 3].
The classification head, $h_{\text{sup}}$, was implemented using three fully connected layers with sizes [512, 256, 10] (where 10 is the number of classes). The same architecture was applied to both classification heads of PointDAN \cite{qin2019pointdan}. The SSL head, $h_{\text{SSL}}$, of DefRec, DAE-point and RS \cite{sauder2019self} was implemented similarly using four 1D convolution layers of sizes [256, 256, 128, 3]. The domain classifier head of DANN \cite{ganin2016domain} was set similar to $h_{\text{sup}}$, namely three fully connected layers of sizes [512, 256, 2]. The reconstruction head of DAE-Global \cite{hassani2019unsupervised} was implemented using four 1D convolution layers of sizes [1024, 1024, 2048, 3072] respectively (where $3072$: $1024 \times 3$ is the reconstruction size). In all heads the nonlinearity was ReLU and a dropout of 0.5 was applied to the first two hidden layers.
Batch normalization \cite{ioffe2015batch} was applied after all convolution layers in $\Phi, h_{\text{sup}}$ and the auxiliary losses.

\begin{table*}[!t]
  \centering
    \scalebox{0.8}{
    \begin{tabular}{p{10em}|p{6em}|p{6em}|p{6em}|p{6em}|p{6em}|p{6em}|p{6em}}
    \toprule
    \textbf{Method} & \textbf{ModelNet to \newline{}ShapeNet} & \textbf{ModelNet to \newline{}ScanNet} & \textbf{ShapeNet to \newline{}ModelNet} & \textbf{ShapeNet to \newline{}ScanNet} & \textbf{ScanNet to\newline{}ModelNet} & \textbf{ScanNet to \newline{}ShapeNet} & \textbf{Avg.}\\
    \midrule
    \midrule
    Unsupervised & 83.3 $\pm$ 0.7 & 43.8 $\pm$ 2.3 & 75.5 $\pm$ 1.8 & 42.5 $\pm$ 1.4 & 63.8 $\pm$ 3.9 & 64.2 $\pm$ 0.8 & 62.2 $\pm$ 1.8\\
    \midrule
    DANN \cite{ganin2016domain}  & 75.3 $\pm$ 0.6 & 41.5 $\pm$ 0.2 & 62.5 $\pm$ 1.4 & 46.1 $\pm$ 2.8 & 53.3 $\pm$ 1.2 & 63.2 $\pm$ 1.2 & 57.0 $\pm$ 1.2\\
    DANN + PCM & 74.8 $\pm$ 2.8 & 42.1 $\pm$ 0.6 & 57.5 $\pm$ 0.4 & 50.9 $\pm$ 1.0 & 43.7 $\pm$ 2.9 & 71.6 $\pm$ 1.0 & 56.8 $\pm$ 1.5\\
    \midrule
    PointDAN \cite{qin2019pointdan} & 82.5 $\pm$ 0.8 & 47.7 $\pm$ 1.0 & 77.0 $\pm$ 0.3 & 48.5 $\pm$ 2.1 & 55.6 $\pm$ 0.6 & 67.2 $\pm$ 2.7 & 63.1 $\pm$ 1.2\\
    PointDAN + PCM & 83.9 $\pm$ 0.3 & 44.8 $\pm$ 1.4 & 63.3 $\pm$ 1.1 & 45.7 $\pm$ 0.7 & 43.6 $\pm$ 2.0 & 56.4 $\pm$ 1.5 & 56.3 $\pm$ 1.2\\
    \midrule
    RS \cite{sauder2019self}  & 81.5 $\pm$ 1.2 & 35.2 $\pm$ 5.9 & 71.9 $\pm$ 1.4 & 39.8 $\pm$ 0.7 & 61.0 $\pm$ 3.3 & 63.6 $\pm$ 3.4 & 58.8 $\pm$ 2.7\\
    RS + PCM & 79.9 $\pm$ 0.8 & 46.7 $\pm$ 4.8 & 75.2 $\pm$ 2.0 & 51.4 $\pm$ 3.9 & 71.8 $\pm$ 2.3 & 71.2 $\pm$ 2.8 & 66.0 $\pm$ 1.6\\
    \midrule
    DAE-Global \cite{hassani2019unsupervised} & 83.5 $\pm$ 0.8 & 42.6 $\pm$ 1.4 & 74.8 $\pm$ 0.8 & 45.5 $\pm$ 1.6 & 64.9 $\pm$ 4.4 & 67.3 $\pm$ 0.6 & 63.1 $\pm$ 1.6\\
    DAE-Global + PCM & 83.1 $\pm$ 0.5 & 47.2 $\pm$ 0.8 & 70.0 $\pm$ 1.0 & 52.8 $\pm$ 0.6 & 67.7 $\pm$ 2.1 & \textbf{73.7 $\pm$ 0.6} & 65.7 $\pm$ 0.9\\
    \midrule
    DAE-Point & 82.5 $\pm$ 0.4 & 40.2 $\pm$ 1.6 & 76.4 $\pm$ 0.7 & 50.2 $\pm$ 0.5 & 66.3 $\pm$ 1.5 & 66.1 $\pm$ 0.5 & 63.6 $\pm$ 0.9\\
    DAE-Point + PCM & \textbf{85.0 $\pm$ 0.5} & 50.2 $\pm$ 1.3 & 74.3 $\pm$ 0.7 & 50.9 $\pm$ 0.8 & 65.1 $\pm$ 1.7 & 72.2 $\pm$ 0.9 & 66.3 $\pm$ 1.0\\
    \midrule
    DefRec (ours) & 83.3 $\pm$ 0.2  & 46.6 $\pm$ 2.0  &\textbf{79.8 $\pm$ 0.5} & 49.9 $\pm$ 1.8 & 70.7 $\pm$ 1.4 & 64.4 $\pm$ 1.2 & 65.8 $\pm$ 1.2\\
    DefRec + PCM (Ours) & 81.7 $\pm$ 0.6 & \textbf{51.8 $\pm$ 0.3} & 78.6 $\pm$ 0.7 & \textbf{54.5 $\pm$ 0.3} & \textbf{73.7 $\pm$ 1.6} & 71.1 $\pm$ 1.4 & \textbf{68.6 $\pm$ 0.8}\\
   \bottomrule
    \end{tabular}%
    }
  \caption{Baselines methods with PCM. Test accuracy on PointDA-10 dataset, averaged over three runs ($\pm$ SEM).}
  \label{table:baselines_with_mixup}%
\end{table*}%

\textbf{Training procedure.} 
In all methods (baselines and ours), during training, we alternate between a batch of source samples and a batch of target samples. We used a fixed batch size of 32 per domain, ADAM optimizer \cite{Kingma2014AdamAM}, and a cosine annealing learning rate scheduler as implemented by PyTorch. We balanced the domains by under-sampling the larger domain, source, or target, in each epoch. We applied grid search over the learning rates \{0.0001, 0.001\} and weight decay \{0.00005, 0.0005\}. In all methods besides PointDAN, we applied a grid search over the auxiliary task weight $\lambda \in$ \{0.25, 1\}. PointDAN has three loss-terms: classification, discrepancy, and MMD. Therefore, for this baseline we applied a grid search over $\{(0.33, 0.33, 0.33), (0.5, 0.25, 0.25)\}$ correspondingly. For DAE-point and DAE-Global we used a Gaussian noise sampled from $\mathcal{N}(0, 0.01)$ as suggested by \cite{hassani2019unsupervised}.
We ran each configuration with 3 different random seeds for 150 epochs and used source-validation-based early stopping. The total training time of DefRec varies between 6-9 hours on a 16g Nvidia V100 GPU, depending on the datasets.

In the paper, we propose three types of deformations to the input point cloud. We implemented these methods with the following settings: 
\begin{itemize}
    \item \textit{Volume-based deformations}. Deformations based on proximity in the input space. We examined two variants of deformations from this type: (a) Split the input space to $k\times k\times k$ equally sized voxels and pick a voxel at random. We tested this method for $k \in$ $\{2, 3\}$ (b) The deformed region is a sphere with a fixed radius $r \in$ $\{0.1, 0.2, ..., 1.0, 2.0\}$ that is centered around one data point selected at random.
    \item \textit{Feature-based deformations}. Deformations based on proximity in the feature space. We examined deformations based on features extracted from layers $1-4$ of the shared feature encoder. The deformed region was set by randomly selecting a point and deforming its $\{100, 150, 200, 300, 500\}$ nearest neighbors in the feature space. 
    \item \textit{Sample-based deformations}. Deformations based on the sampling direction. For the gradient and the Lambertian methods, we followed the protocol suggested by \cite{hermosilla2018monte}. For the split method, we randomly selected a cut off according to a beta distribution with parameters $a=2.0, b=5.0$. 
\end{itemize}

\subsection{PointSegDA dataset} \label{sec:pointSegda_imp}
Similar to the classification case, during training we applied jittering with standard deviation and clip parameters of 0.01 and 0.02 respectively, and random rotations to shapes about the $Z$ axis only. The training procedure and network architecture were similar to the ones described in Section \ref{sec:pointda_imp} with the following exceptions:
\begin{itemize}
    \item \textit{Network}. We used the DGCNN feature extractor and segmentation head for segmentation tasks. Unlike the classification head, the segmentation head takes the global feature vector concatenated to feature representations of the points. It was implemented using four 1D convolution layers of sizes [256, 256, 128, 8], where 8 is the number of classes.
    \item \textit{Training procedure}. The batch size was set to 16 per domain, the number of epochs was set to 200, and a grid search over the auxiliary task weight $\lambda$ was done in $\{0.05, 0.1, 0.2\}$. Multi-level Adapt-SegMap \cite{tsai2018learning} was implemented with two segmentation heads and two discriminators, all having the architecture described in the previous item. For Adapt-SegMap we applied grid search over the adversarial tasks weights in $\{(0.0002, 0.001), (0.0002, 0.01), (0.002, 0.001),\\(0.002, 0.01)\}$
    \item \textit{DefRec hyperparameters}. (i) Volume-based deformations: we searched over the hyperparameters $k = 3$ and $r \in \{0.2, 0.3, 0.4, 0.5\}$. (ii) Feature-based deformations: we considered only the layers $2-3$. The deformed region was set by randomly selecting a point and deforming its $\{400, 600\}$ nearest neighbors in the feature space.
\end{itemize}

\begin{table*}[!t]
  \centering
  \scalebox{0.8}{
    \begin{tabular}{p{10em}|p{6em}|p{6em}|p{6em}|p{6em}|p{6em}|p{6em}|p{5em}}
    \toprule
    \textbf{Method} & \textbf{ModelNet to \newline{}ShapeNet} & \textbf{ModelNet to \newline{}ScanNet} & \textbf{ShapeNet to \newline{}ModelNet} & \textbf{ShapeNet to \newline{}ScanNet} & \textbf{ScanNet to\newline{}ModelNet} & \textbf{ScanNet to \newline{}ShapeNet} & \textbf{Avg.}\\
    \midrule
    DefRec & 84.7 $\pm$ 0.5 & 44.3 $\pm$ 2.3 & 79.3 $\pm$ 0.9 & 49.7 $\pm$ 1.0 & 66.3 $\pm$ 1.6 & 68.1 $\pm$ 0.9 & 65.4 $\pm$ 1.2\\
    \midrule
    DefRec + PCM & 84.0 $\pm$ 0.3 & 55.0 $\pm$ 1.2 & 74.7 $\pm$ 1.0 & 54.4 $\pm$ 0.1 & 69.7 $\pm$ 0.6 & 76.2 $\pm$ 0.3 & 69.0 $\pm$ 0.6\\
    \bottomrule
    \end{tabular}%
    }
  \caption{
  Combining deformation strategies. Test accuracy on PointDA-10 dataset, averaged over three runs ($\pm$ SEM).}
  \label{table:mixed_def}%
\end{table*}%

\section{Additional experiments} \label{sec:additional_exp}
\subsection{PCM on baselines}  \label{sec:baselines_with_pcm}
Table~\ref{table:baselines_with_mixup} compares DefRec + PCM to the baseline methods combined with the PCM module.
From the table, we notice that PCM boosts the performance of RS, DAE-Global, and DAE-Point but less so for DANN and PointDAN. Nevertheless, our proposed approach of combining PCM with DefRec is still superior. PointDAN uses a discrepancy loss which entails having two classification heads. Therefore, we speculate that adding PCM in this scenario hurts the performance. Combining PCM with several classification heads is an interesting research direction which we leave for future work.

\subsection{Combining deformation strategies} \label{sec:mixing_def}
DefRec procedure requires first to choose the deformation type and then hyperparameters specific for the type. This process may be cumbersome. Therefore, here we suggest an alternative protocol. Instead of choosing a specific deformation type, we propose to apply all of them with equal weight. For efficiency, this is achieved by choosing each deformation with a probability of $1/3$ in each batch. The hyperparameters of each type of deformation were set according to a sensitivity analysis based on the source-validation accuracy. As expected, we found the chosen hyperparameters to be highly correlated with the ones presented in Fig. \ref{fig:Analysis_families} (e.g., radius of 0.2 for the volume-based). Table \ref{table:mixed_def} shows the results of applying this protocol. Comparing these results with the ones in Table \ref{table:main_table} shows that these two alternatives are comparable. On some adaptations there is a significant improvement, for example, in \textit{ModelNet to ShapeNet} and \textit{ModelNet to ScanNet}, when applying PCM, the accuracy increase by 2\% and 3\% respectively.

\section{Estimating target perplexity} \label{sec:perplexity}
A key property of a DA solution is the ability to find an alignment between source and target distributions that is also discriminative \cite{saito2018maximum}. To test that we suggest measuring the log perplexity of target test data representation under a model fitted by source test data representation. Here we consider the representation of samples as the activations of the last hidden layer in the classification network. The log perplexity measures the average number of bits required to encode a test sample. A lower value indicates a better model with less uncertainty in it.

Let  $(x^t_1,y^t_1), ..., (x^t_1,y^t_n) \in T$ be a set of target instances. We note by $n_c$ the number of target instances from class $c$. Using the chain rule, the likelihood of the joint distribution $p(x^t_j,y^t_j = c)$ can be estimated by finding $P(x^t_j|y^t_j = c)$ and $P(y^t_j = c)$. To model $P(x^t_j|y^t_j = c)$ we fit a Gaussian distribution $ N(\mu{_c}, \Sigma{_c})$ based on source samples from class $c$ using maximum likelihood. To model $p(y^t_j = c)$ we take the proportion of source samples in class $c$.

Modeling the class conditional distribution with a Gaussian distribution relates to the notion proposed in \cite{snell2017prototypical}. \cite{snell2017prototypical} suggested to represent each class with a prototype (the mean embeddings of samples belonging to the class) and assign a new instance to the class associated with the closest prototype. The distance metric used is the squared Euclidean distance. This method is equivalent to fitting a Gaussian distribution for each class with a unit covariance matrix.

\begin{table}[!b]
  \centering
    \begin{tabular}{l|l|l}
    \toprule
    \multicolumn{1}{p{6em}|}{Method} & \multicolumn{1}{p{4em}|}{Standrad \newline{}Perplexity} & \multicolumn{1}{p{7em}}{Class-Balanced \newline{}Perplexity} \\
    \midrule
    \multicolumn{3}{c}{ModelNet to ScanNet} \\
    \midrule
    PointDAN & \textbf{25.3 $\pm$ 1.4} & 36.4 $\pm$ 1.1 \\
    \midrule
    DefRec + PCM & 29.8 $\pm$ 1.9 & \textbf{33.1 $\pm$ 1.4} \\
    \midrule
    \multicolumn{3}{c}{ModelNet to ShapeNet} \\
    \midrule
    PointDAN & 6.8 $\pm$ 0.4 & 23.6 $\pm$ 3.5 \\
    \midrule
    DefRec + PCM & \textbf{6.1 $\pm$ 0.3} & \textbf{20.4 $\pm$ 3.0} \\
    \bottomrule
    \end{tabular}%
  \caption{Log perplexity ($\pm$ SEM). Lower is better.}
  \label{perplexity}%
\end{table}%

The log perplexity of the target is (noted as standard perplexity here after):
\begin{equation}
    L(T) = \sum_{c=1}^{10}\sum_{j=1}^{n_c} \frac{1}{n} \log \left(p(x^t_j|y^t_j = c) p(y^t_j = c) \right)
\label{dist}
\end{equation}

Alternatively we can measure the mean of a class-balanced log perplexity (noted as class-balanced perplexity here after):
\begin{equation} 
L(T) = \frac{1}{10} \sum_{c=1}^{10}\sum_{j=1}^{n_c} \frac{1}{n_c} \log \left(p(x^t_j|y^t_j = c) p(y^t_j = c) \right)
\end{equation}

\begin{figure*}[!t]
    \centering
    \subfloat[DefRec + PCM]{
        \includegraphics[width=0.43\linewidth]{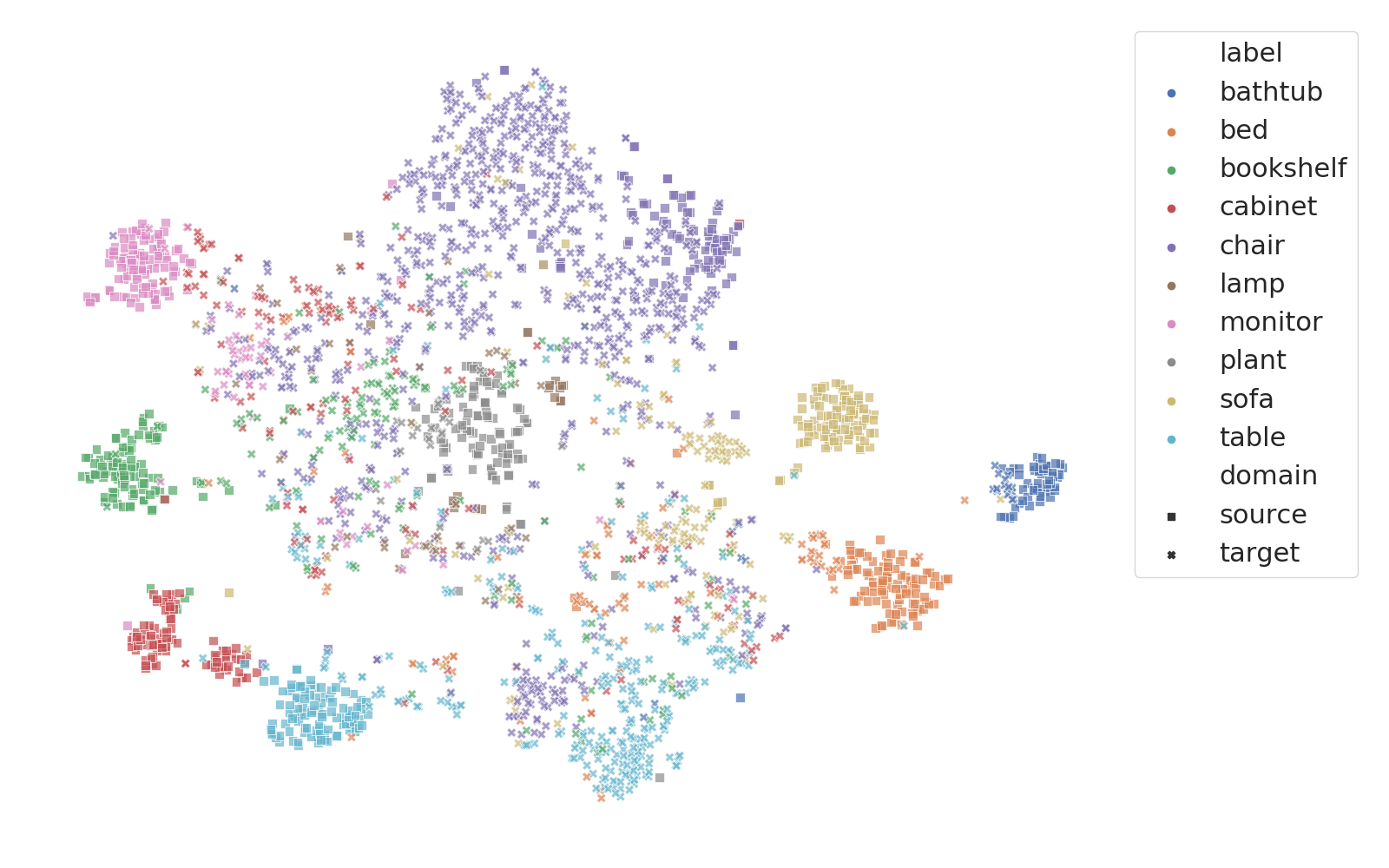}
        \label{tsne_cls:1a}        
    }
    \subfloat[PointDAN]{
        \includegraphics[width=0.43\linewidth]{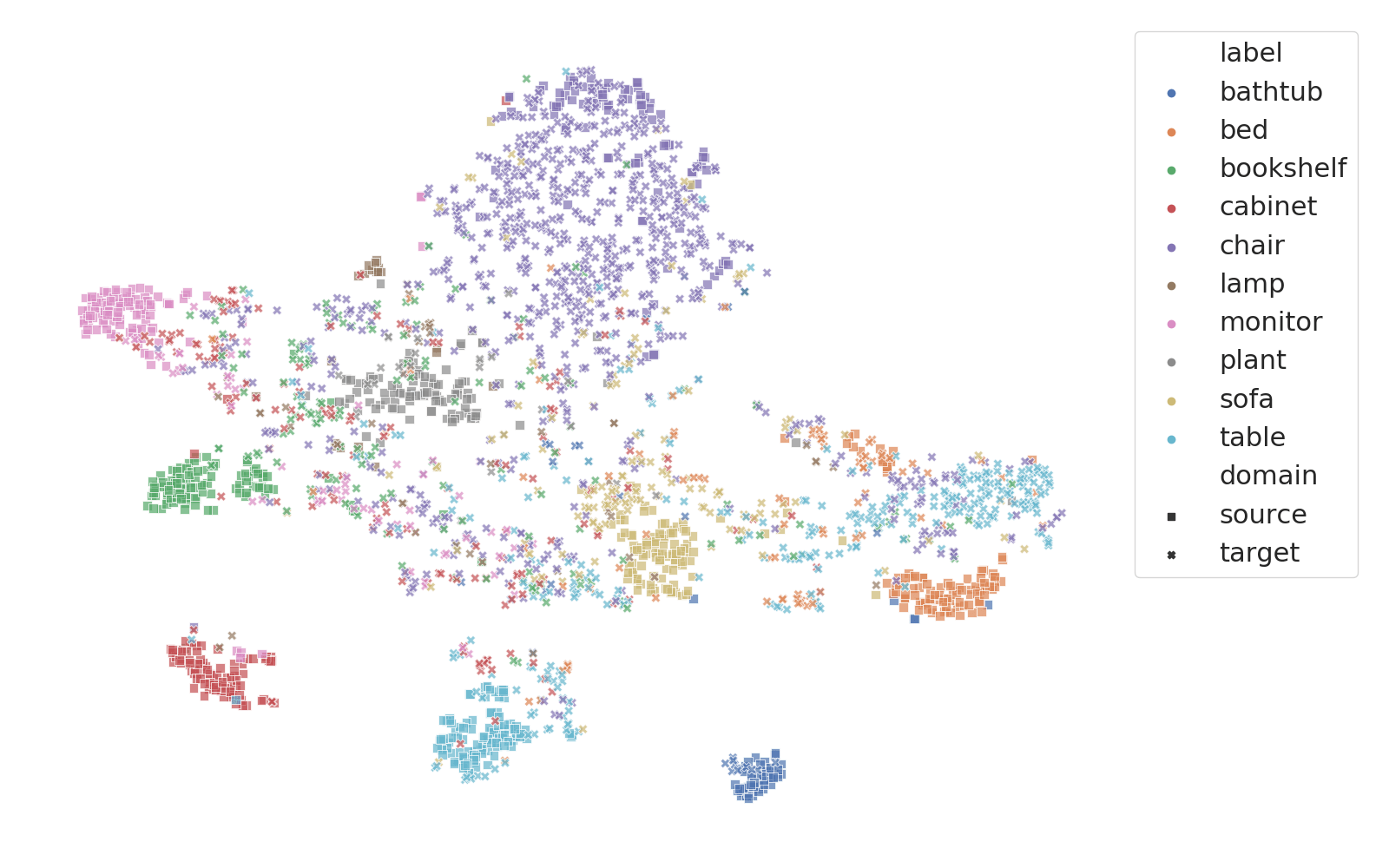}
        \label{fig_cls:1b}
    }
    \caption{The distribution of samples for the adaptation ModelNet to ScanNet.}
    \label{fig:tsne_cls_mn_to_sn}
\end{figure*}

\begin{figure*}[!t]
    \centering
    \subfloat[DefRec + PCM]{
        \includegraphics[width=0.43\linewidth]{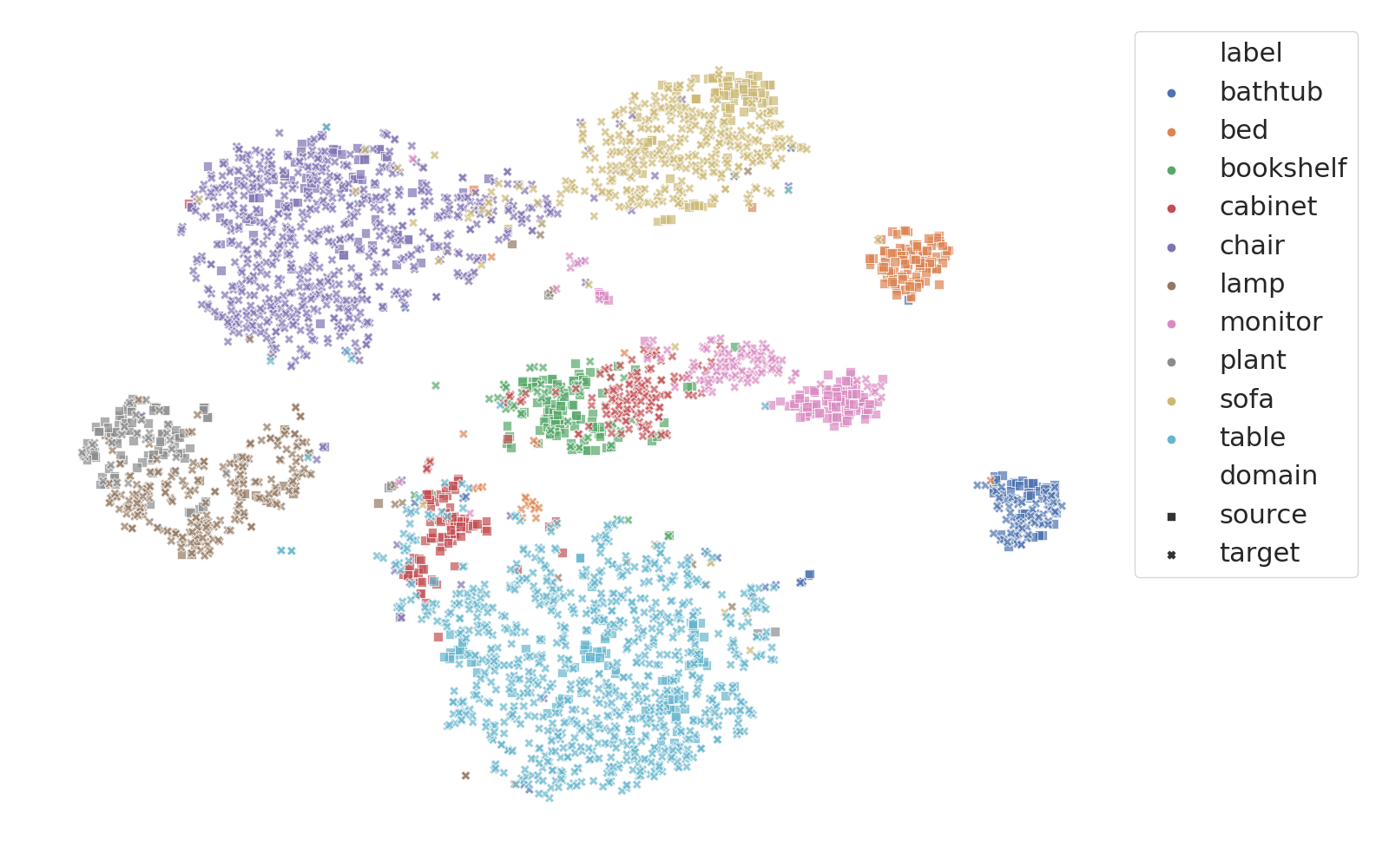}
        \label{tsne_cls:2a}        
    }
    \subfloat[PointDAN]{
        \includegraphics[width=0.43\linewidth]{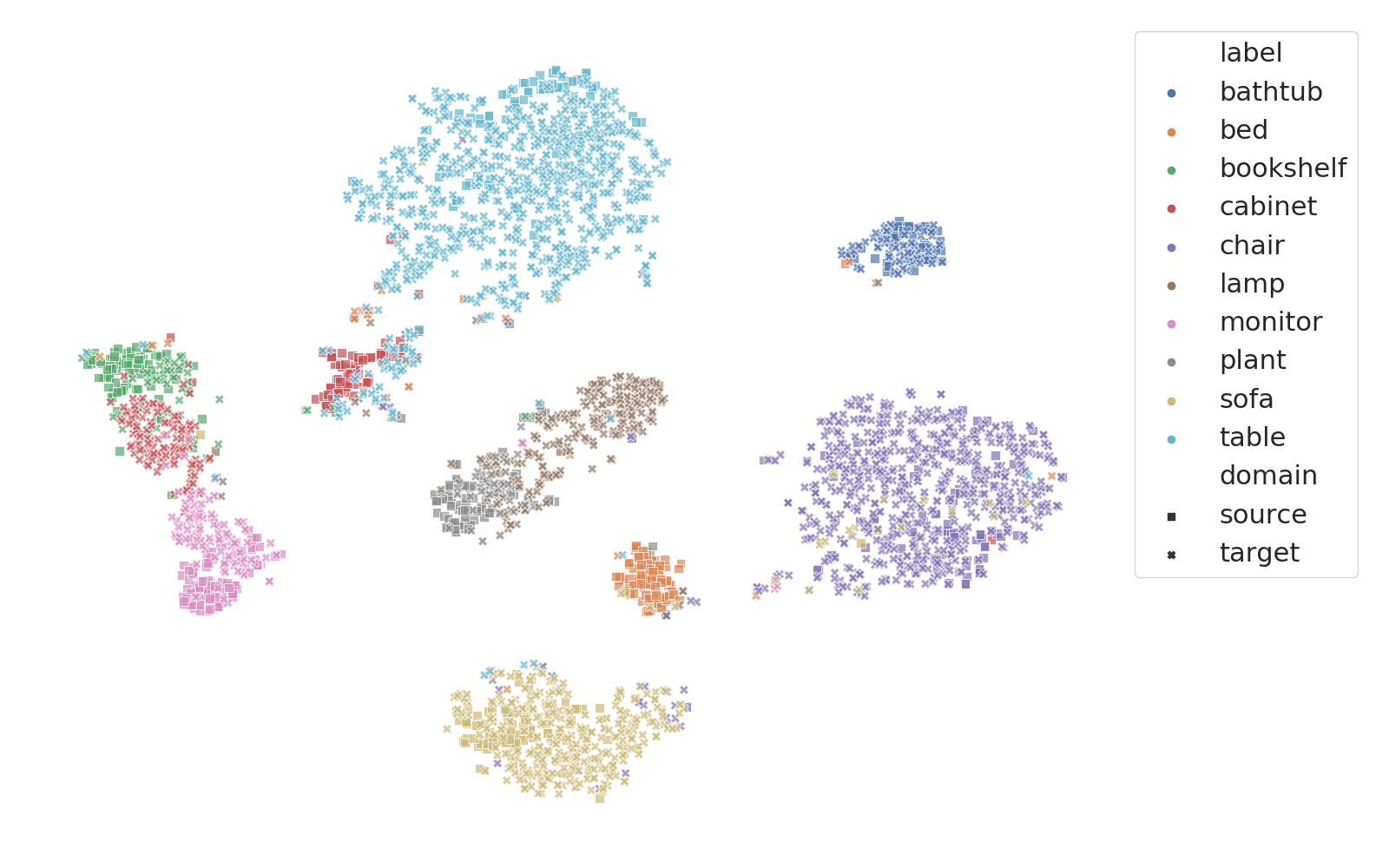}
        \label{fig_cls:2b}
    }
    \caption{The distribution of samples for the adaptation ModelNet to ShapeNet.}
    \label{fig:tsne_cls_mn_to_sh}
\end{figure*}

Table~\ref{perplexity} shows the standard perplexity and class-balanced perplexity of DefRec + PCM of our best model that was chosen based on the source-validation set and PointDAN \cite{qin2019pointdan} for the adaptations \textit{ModelNet to ScanNet} and \textit{ModelNet to ShapeNet}. Estimating the perplexity on the original space requires estimating a covariance matrix from a relatively small number of samples which results in a degenerate matrix. Therefore, we estimated the perplexity after applying dimensionality reduction to a 2D space using t-SNE. We ran t-SNE with the same configurations with ten different seeds and reported the mean and standard error of the mean. In Figures \ref{fig:tsne_cls_mn_to_sn} and \ref{fig:tsne_cls_mn_to_sh} we plot the t-SNE representations of one of the seeds. 

From the table and the figures, we see that our method creates target and source representations that are more similar. In both adaptations, the class-balanced perplexity of our model is smaller. This is an indication that our model is doing a better job at learning under-represented classes. We note that PointDAN creates a denser representation of some classes (especially well-represented classes such as \textit{Chair} and \textit{Table}) however, they are not mixed better between source and target. 

\section{Shape reconstruction} \label{sec:shape_rec_supp}
\begin{figure*}[!t]
    \centering
    \includegraphics[width=\textwidth, height=20cm]{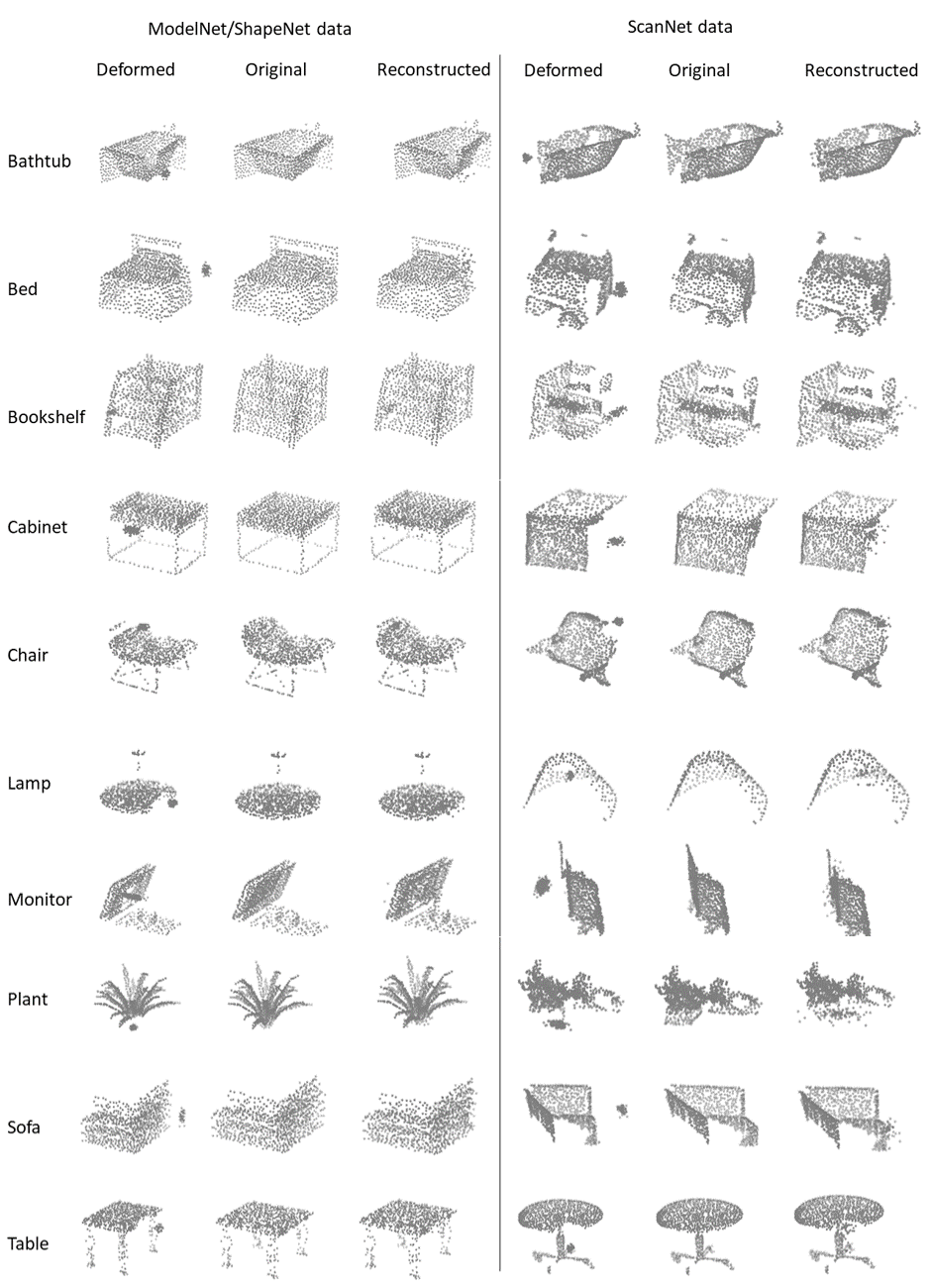}
    \caption{Illustration of target reconstruction of all classes. Each triplet shows a sample deformed using DefRec, the ground truth original, and the resulting reconstruction. Left triplets: ShapeNet/ModelNet. Right triplets: ScanNet.}
    \label{fig:Reconstruction}
\end{figure*}

\begin{figure*}[!t]
    \centering
    \includegraphics[width=0.35\linewidth, height=4cm]{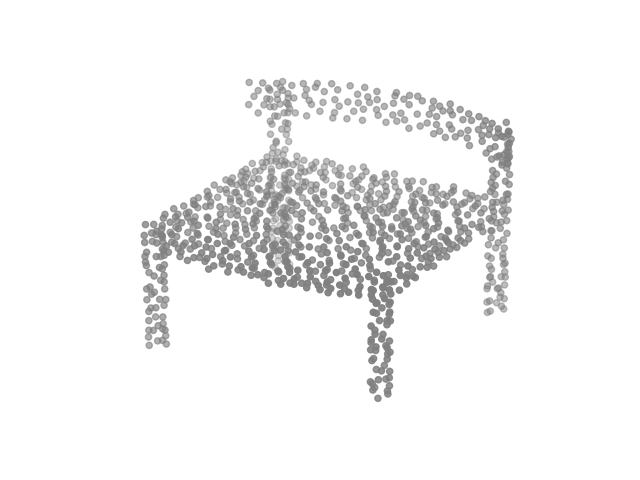}
    \vfill
    \includegraphics[width=0.3\linewidth, height=3cm]{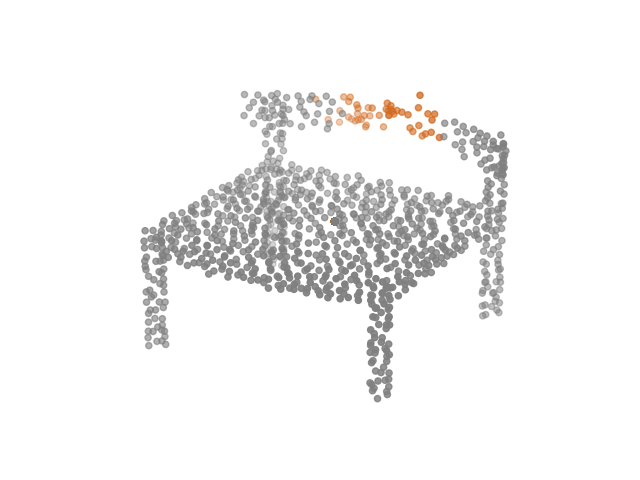}
    \includegraphics[width=0.3\linewidth, height=3cm]{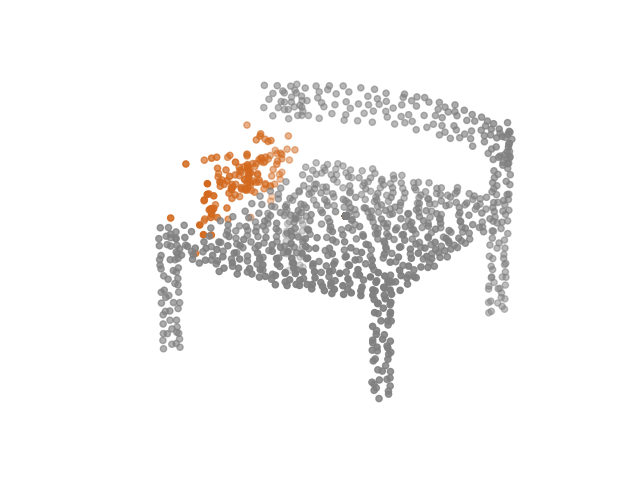}
    \includegraphics[width=0.3\linewidth, height=3cm]{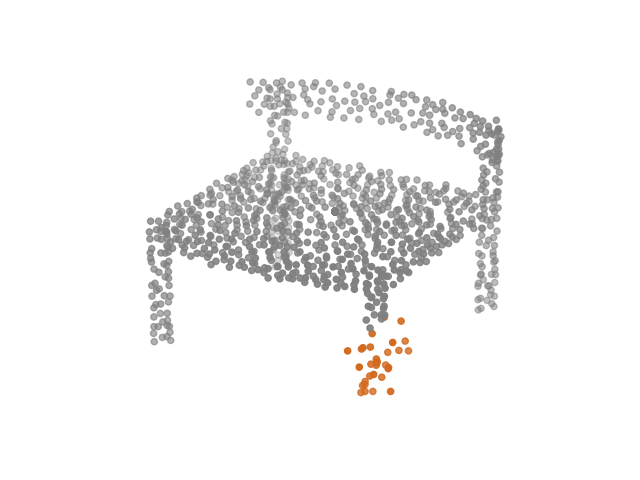}
    \vfill
    \includegraphics[width=0.3\linewidth, height=3cm]{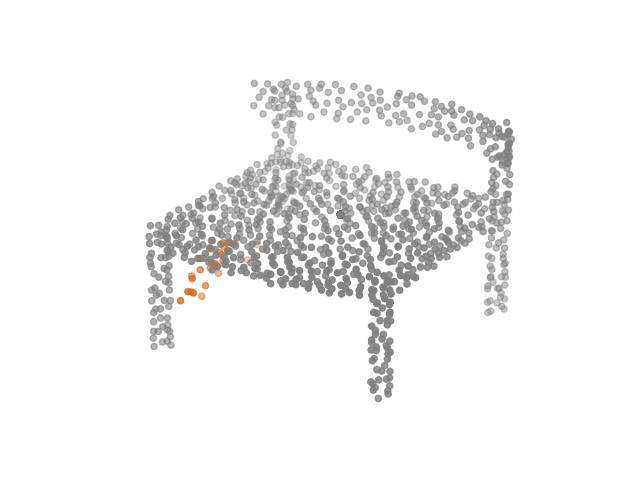}
    \includegraphics[width=0.3\linewidth, height=3cm]{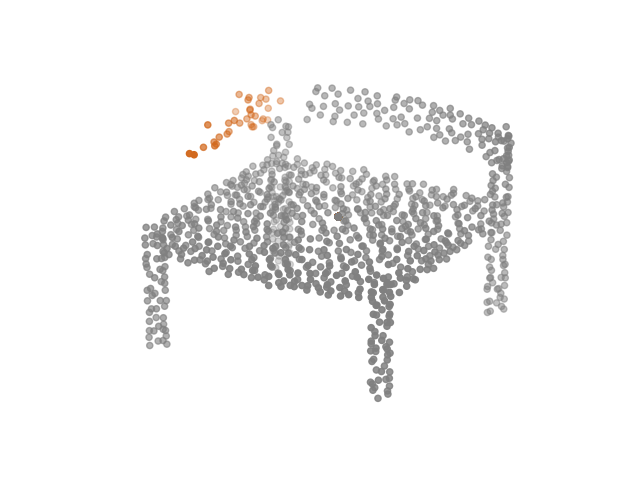}
    \includegraphics[width=0.3\linewidth, height=3cm]{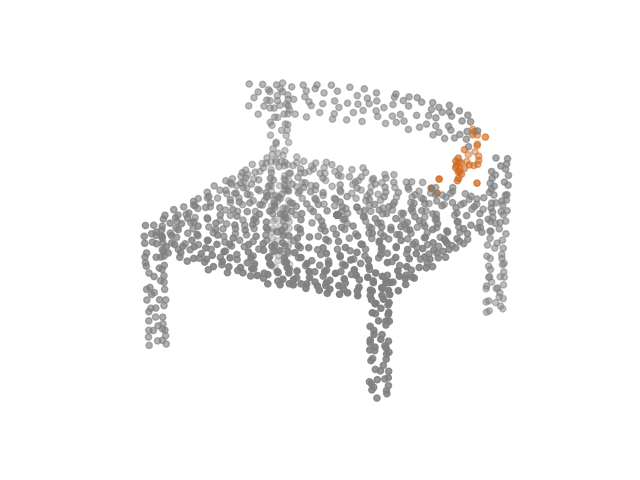}
    \vfill
    \includegraphics[width=0.3\linewidth, height=3cm]{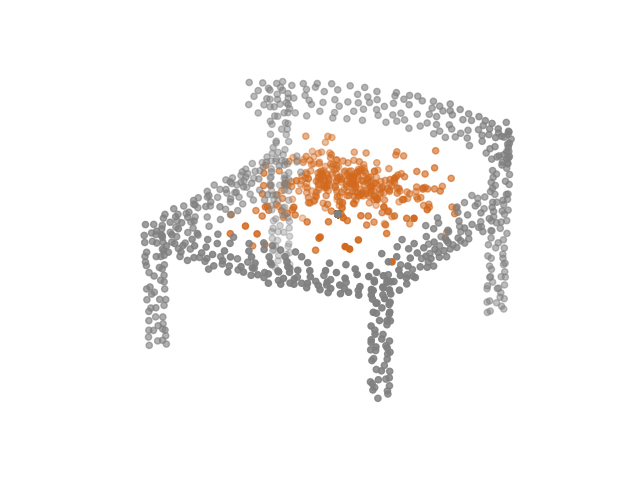}
    \includegraphics[width=0.3\linewidth, height=3cm]{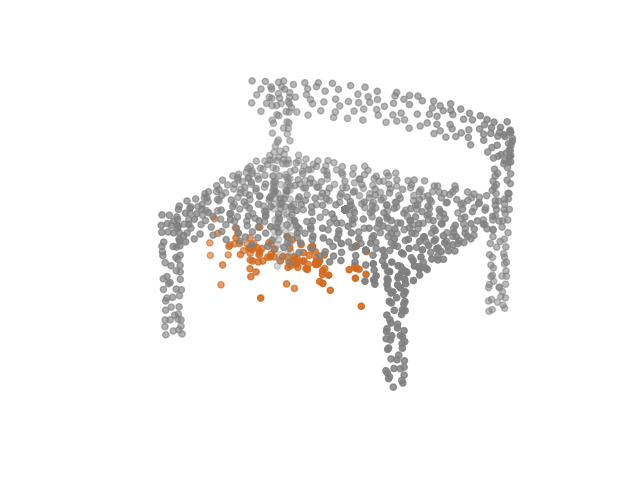}
    \includegraphics[width=0.3\linewidth, height=3cm]{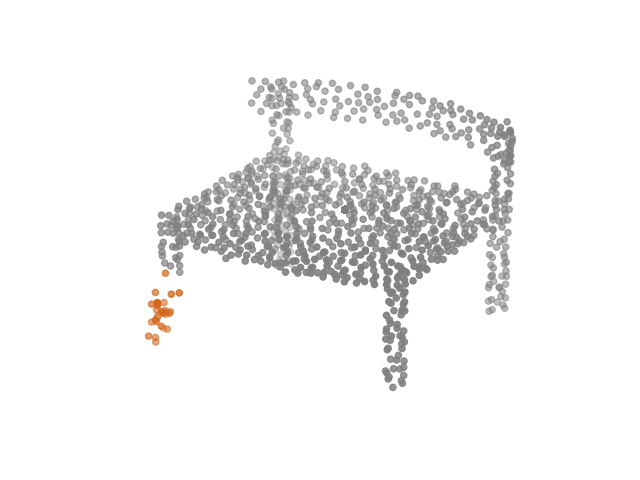}
    \vfill
    \includegraphics[width=0.3\linewidth, height=3cm]{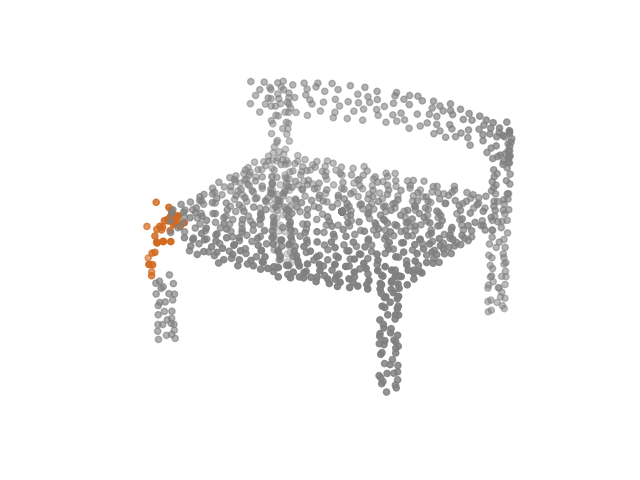}
    \includegraphics[width=0.3\linewidth, height=3cm]{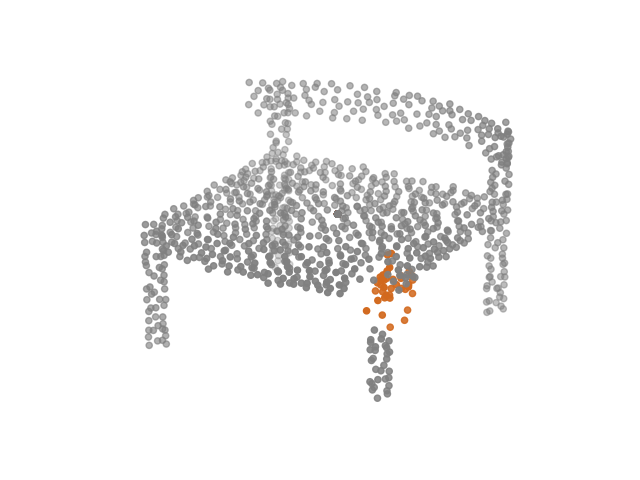}
    \includegraphics[width=0.3\linewidth, height=3cm]{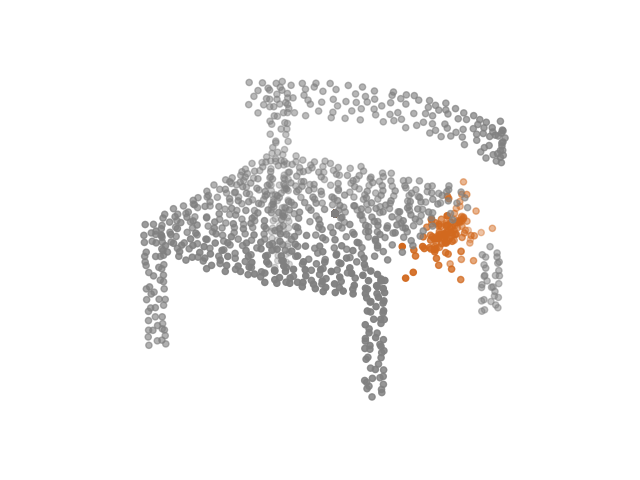}
    \vfill
    \includegraphics[width=0.3\linewidth, height=3cm]{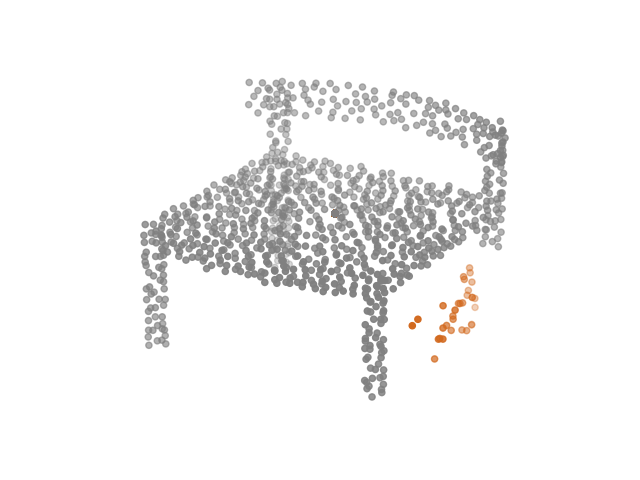}
    \includegraphics[width=0.3\linewidth, height=3cm]{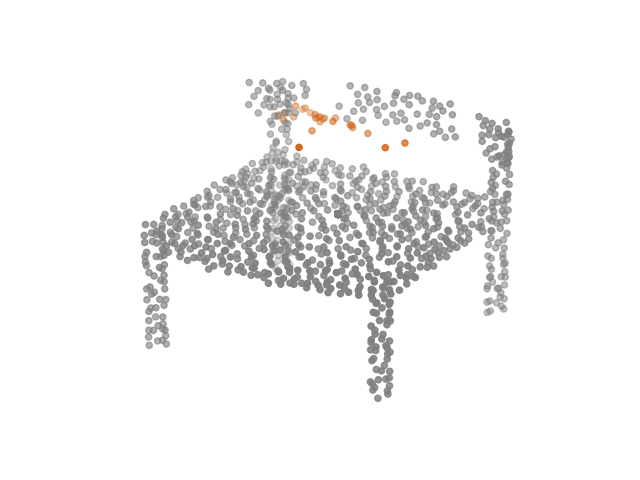}
    \caption{Reconstruction of a chair object from deformation of different regions in it by DefRec. The object in the first row is the ground truth. Below it are the reconstructed shapes, each with a deformation of different region in the object. Reconstructed region is marked by orange.}
    \label{fig:chair_rec}
\end{figure*}

\begin{figure*}[!t]
    \centering
    \includegraphics[width=0.35\linewidth, height=4cm]{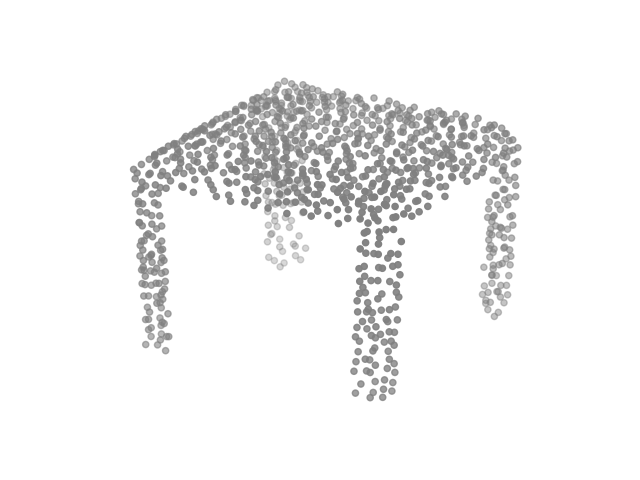}
    \vfill
    \includegraphics[width=0.3\linewidth, height=3cm]{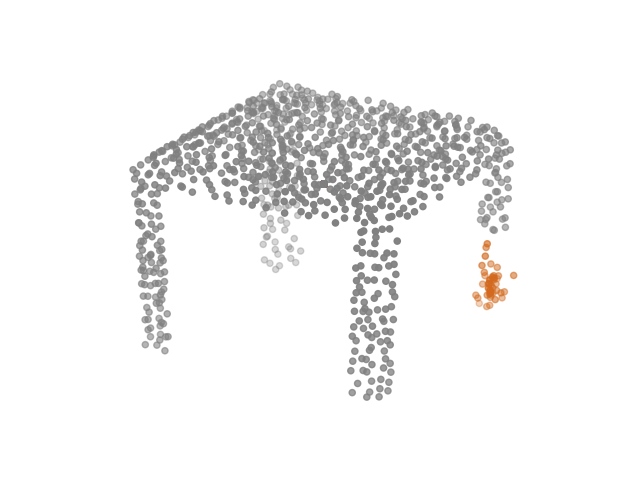}
    \includegraphics[width=0.3\linewidth, height=3cm]{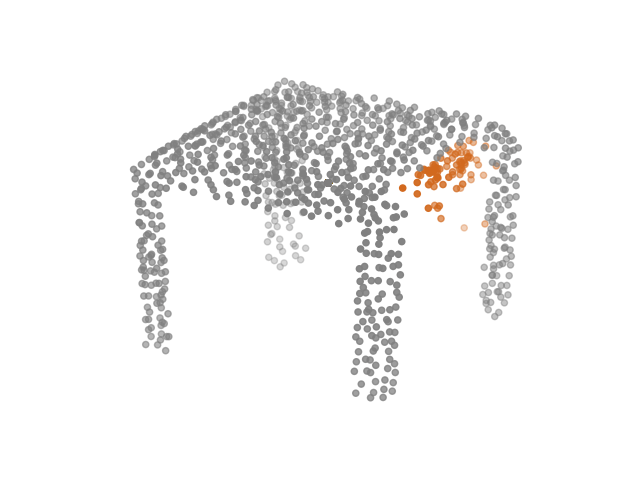}
    \includegraphics[width=0.3\linewidth, height=3cm]{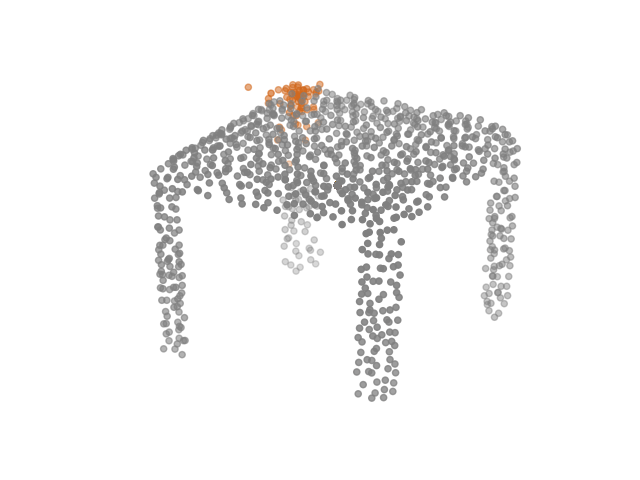}
    \vfill
    \includegraphics[width=0.3\linewidth, height=3cm]{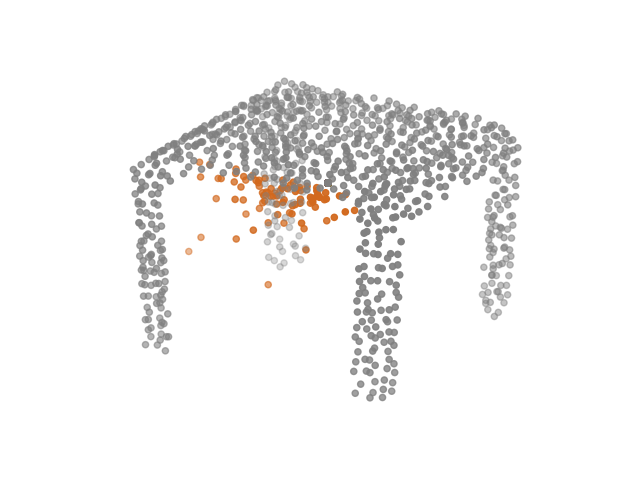}
    \includegraphics[width=0.3\linewidth, height=3cm]{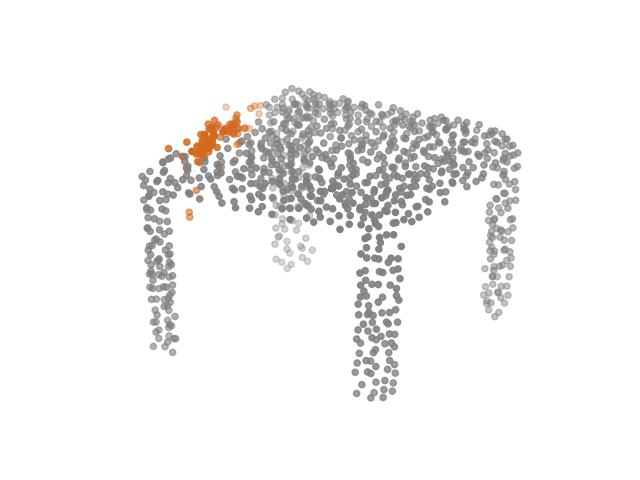}
    \includegraphics[width=0.3\linewidth, height=3cm]{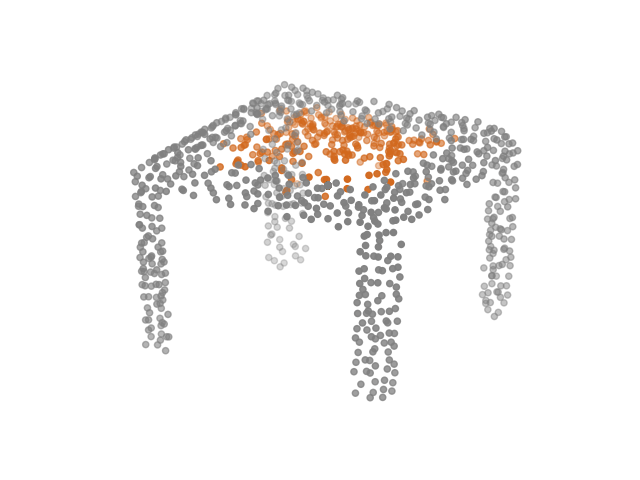}
    \vfill
    \includegraphics[width=0.3\linewidth, height=3cm]{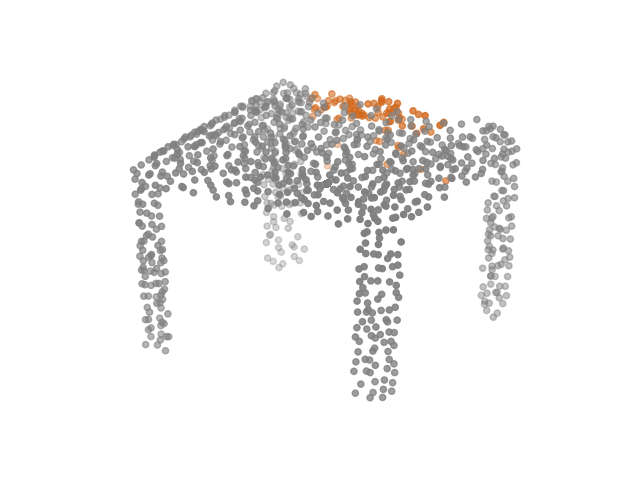}
    \includegraphics[width=0.3\linewidth, height=3cm]{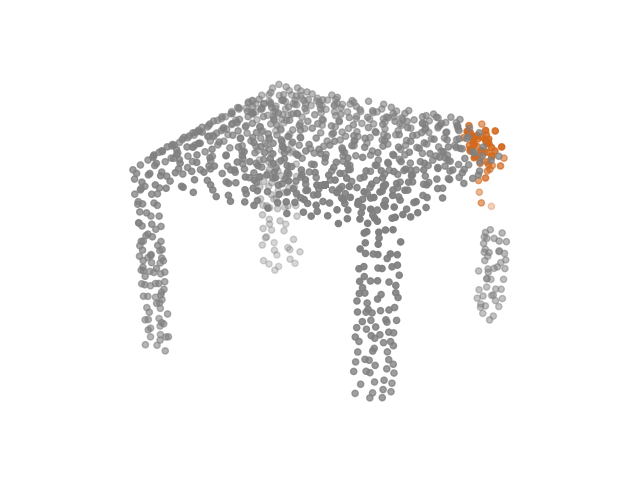}
    \includegraphics[width=0.3\linewidth, height=3cm]{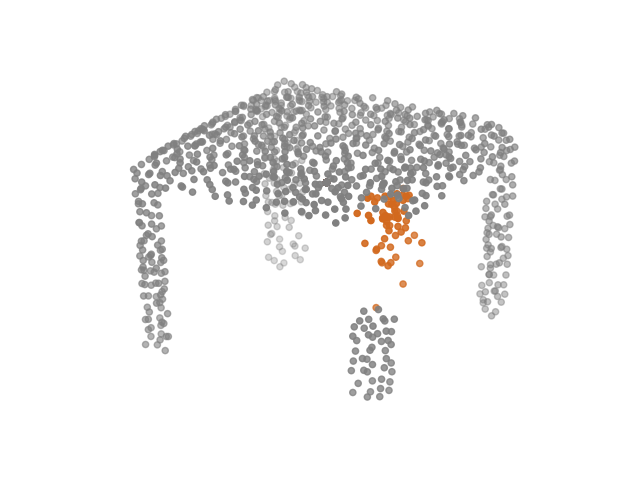}
    \vfill
    \includegraphics[width=0.3\linewidth, height=3cm]{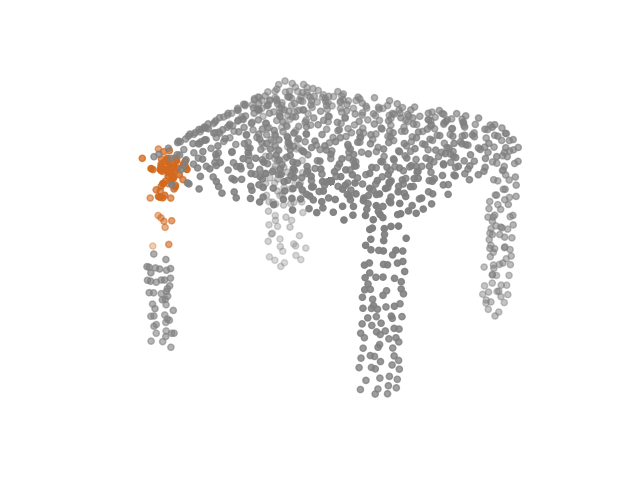}
    \includegraphics[width=0.3\linewidth, height=3cm]{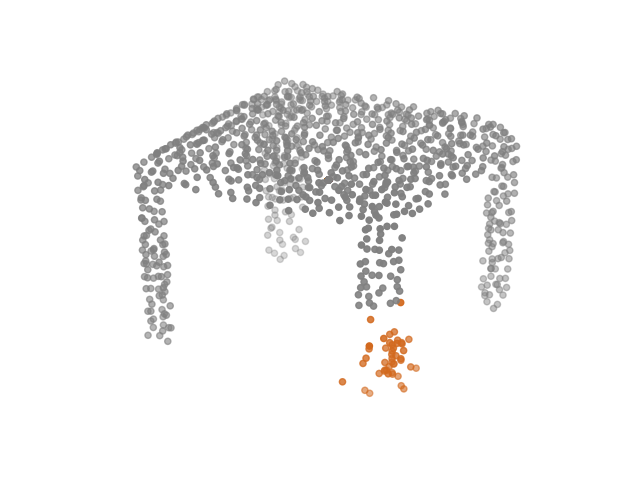}
    \includegraphics[width=0.3\linewidth, height=3cm]{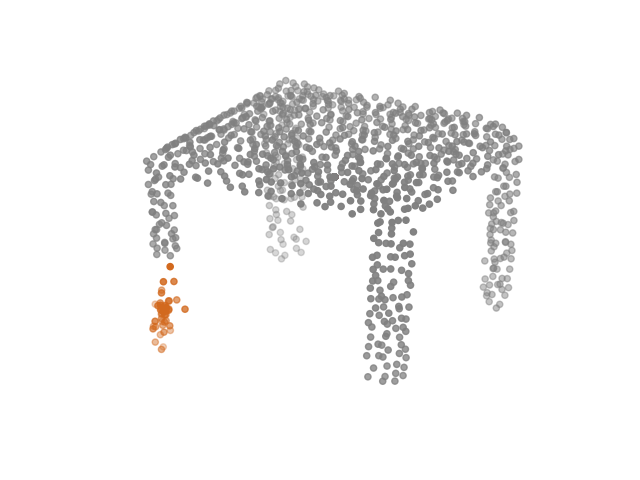}
    \vfill
    \includegraphics[width=0.3\linewidth]{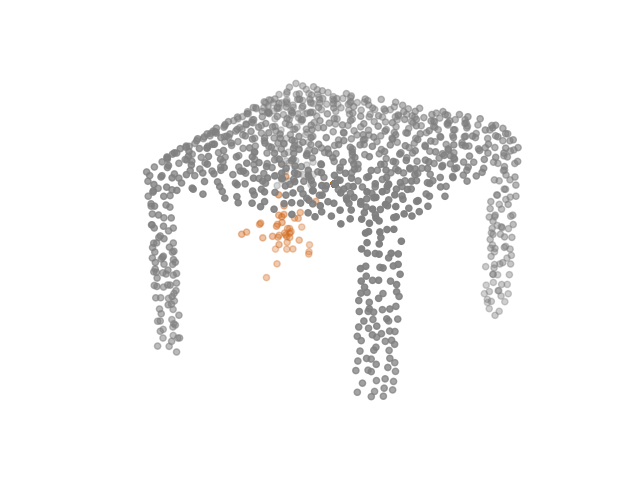}
    \caption{Reconstruction of a Table object from deformation of different regions in it by DefRec. The object in the first row is the ground truth. Below it are the reconstructed shapes, each with a deformation of different region in the object. Reconstructed region is marked by orange.}
    \label{fig:table_rec}
\end{figure*}

Although we developed DefRec for the purpose of DA we expect it to learn reasonable reconstructions from point cloud deformations. Figures~\ref{fig:Reconstruction}-\ref{fig:table_rec} show 
DefRec reconstruction of deformed shapes by the first variant of the volume-based type. Namely, we split the input space to $3\times 3\times 3$ voxels and pick one voxel uniformly at random. 

Figure \ref{fig:Reconstruction} demonstrate DefRec reconstruction of a shapes from all classes in the data for the simulated domains (left column) and the real domain (right column). Images of the same object are presented in the following order from left to right: the deformed shape (the input to the network), the original shape (the ground truth) and the reconstructed shape by the network. From the figure, it seems that the network manages to learn two important things: (1) It learns to recognize the deformed region and  (2) it learns to reconstruct the region in a way that preserves the original shape. Note how in some cases, such as \textit{Monitor} on the left column and \textit{Lamp} on the right column, the reconstruction is not entirely consistent with the ground truth. The network reconstructs the object in a different (but still plausible) manner.

Figures \ref{fig:chair_rec} and \ref{fig:table_rec} show DefRec reconstruction of \textit{Chair} and \textit{Table} objects respectively from deformations of different voxels in the objects. It can be seen that the network learns to reconstruct some regions nicely (such as the chair's top rail or table legs) while it fails to reconstruct well other regions (such as the chair's seat).

\end{document}